\documentclass{article}

\usepackage[final]{corl_2019} 

\usepackage[utf8]{inputenc} 
\usepackage[T1]{fontenc}    
\usepackage[english]{babel}
\usepackage{hyperref}       
\usepackage{url}            
\usepackage{booktabs}       
\usepackage{amsfonts}       
\usepackage{nicefrac}       
\usepackage{microtype}      

\usepackage{amsmath}
\usepackage{amssymb}
\usepackage{mathabx}
\usepackage{bm}
\usepackage{graphicx}

\usepackage[export]{adjustbox}
\usepackage{subfig}
\usepackage[noend]{algpseudocode} 
\usepackage{algorithm}
\usepackage{multirow}
\usepackage{array}
\usepackage{verbatim}

\title{Teacher algorithms for curriculum learning of \mbox{Deep RL} in continuously parameterized environments}

\author{
  R\'emy Portelas\\
  Inria (FR) \\
  \And
  C\'edric Colas\\
  Inria (FR) \\
  \And
  Katja Hofmann\\
  Microsoft Research (UK) \\
  \And
  Pierre-Yves Oudeyer\\
  Inria (FR) \\
}

\begin{document}
\maketitle

\begin{abstract}

       
      We consider the problem of how a teacher algorithm can enable an unknown Deep Reinforcement Learning (DRL) student to become good at a skill over a wide range of diverse environments. To do so, we study how a teacher algorithm can learn to generate a learning curriculum, whereby it sequentially samples parameters controlling a stochastic procedural generation of environments. Because it does not initially know the capacities of its student, a key challenge for the teacher is to discover which environments are easy, difficult or unlearnable, and in what order to propose them to maximize the efficiency of learning over the learnable ones. To achieve this, this problem is transformed into a surrogate continuous bandit problem where the teacher samples environments in order to maximize absolute learning progress of its student. We present a new algorithm modeling absolute learning progress with Gaussian mixture models (ALP-GMM). We also adapt existing algorithms and provide a complete study in the context of DRL. Using parameterized variants of the BipedalWalker environment, we study their efficiency to personalize a learning curriculum for different learners (embodiments), their robustness to the ratio of learnable/unlearnable environments, and their scalability to non-linear and high-dimensional parameter spaces. Videos and code are available at \small{\url{https://github.com/flowersteam/teachDeepRL}}.
    
\end{abstract}

\keywords{Deep Reinforcement Learning, Teacher-Student Learning, Curriculum Learning, Learning Progress, Curiosity, Parameterized Procedural Environments}

\section{Introduction}


We address the \textit{strategic student problem}. This problem is well known in the developmental robotics community \citep{strategicstudent}, and formalizes a setting where an agent has to sequentially select tasks to train on to maximize its average competence over the whole set of tasks after a given number of interactions. To address this problem, several works \citep{playgroundexp,baranes2013active, moulinfriergmm} proposed to use automated Curriculum Learning (CL) strategies based on Learning Progress (LP) \citep{devrobbook}, and showed that population-based algorithms can benefit from such techniques. Inspired by these initial results, similar approaches \citep{curious} were then successfully applied to DRL agents in continuous control scenarios with discrete sets of \textit{goals}, here defined as tasks varying only by their reward functions (e.g reaching various target positions in a maze). Promising results  were also observed when learning to navigate in discrete sets of \textit{environments}, defined as tasks differing by their state space (e.g escaping from a set of mazes) \citep{tscl, tscllike}.



In this paper, we study for the first time whether LP-based curriculum learning methods are able to scaffold generalist DRL agents in continuously parameterized environments. We compare the reuse of Robust Intelligent Adaptive Curiosity (RIAC) \citep{riac} in this new context to Absolute Learning Progress - Gaussian Mixture Model (ALP-GMM), a new GMM-based approach inspired by earlier work on developmental robotics \citep{moulinfriergmm} that is well suited for DRL agents. Both these methods rely on Absolute Learning Progress (ALP) as a surrogate objective to optimize with the aim to maximize average competence over a given parameter space. Importantly, our approaches do not assume a direct mapping from parameters to environments, meaning that a given parameter vector encodes a distribution of environments with similar properties, which is closer to real-world scenarios where stochasticity is an issue.

Recent work \citep{poet} already showed impressive results in continuously parameterized environments. The POET approach proved itself to be capable of generating and mastering a large set of diverse BipedalWalker environments. However, their work differs from ours as they evolve a population of agents where each individual agent is specialized for a single specific deterministic environment, whereas we seek to scaffold the learning of a single generalist agent in a training regime where it never sees the same exact environment twice.


As our approaches make few assumptions, they can deal with ill-defined parameter spaces that include unfeasible subspaces and irrelevant parameter dimensions. This makes them particularly well suited to complex continuous parameter spaces in which expert-knowledge is difficult to acquire. We formulate the Continuous Teacher-Student (CTS) framework to cover this scope of challenges, opening the range of potential applications.


\textbf{Main contributions:}
\begin{itemize}
    \item A Continuous Teacher-Student setup enabling to frame Teacher-Student interactions for ill-defined continuous parameter spaces encoding distributions of tasks. See Sec.  \ref{sec:framework}.
    \item Design of two parameterized BipedalWalker environments, well-suited to benchmark CL approaches on continuous parameter spaces encoding distributions of environments with procedural generation. See Sec. \ref{ssec:PBW}.
    \item ALP-GMM, a CL approach based on Gaussian Model Mixture and absolute LP that is well suited for DRL agents learning continuously parameterized tasks. See Sec. \ref{ssec:alpalgos}.
    \item First study of ALP-based teacher algorithms leveraged to scaffold the learning of generalist DRL agents in continuously parameterized environments. See Sec. \ref{sec:results}.
    
\end{itemize}

\section{Related work}

\textit{Curriculum learning}, as formulated in the supervised machine learning community, initially refers to techniques aimed at organizing labeled data to optimize the training of neural networks \citep{elman,krueger,bengiocl}. Concurrently, the RL community has been experimenting with transfer learning methods, providing ways to improve an agent on a target task by pre-training on an easier source task \citep{taylortransfer}. These two lines of work were combined and gave birth to curriculum learning for RL, that is, methods organizing the order in which tasks are presented to a learning agent so as to maximize its performance on one or several target tasks.

Learning progress has often been used as an intrinsically motivated objective to automate curriculum learning in developmental robotics \citep{devrobbook}, leading to successful applications in population-based robotic control in simulated \citep{moulinfriergmm,modularforestier} and real-world environments \citep{playgroundexp, baranes2013active}. LP was also used to accelerate the training of LSTMs and neural turing machines \citep{graves2017automated}, and to personalize sequences of exercises for children in educational technologies \citep{zpdes}.


A similar Teacher-Student framework was proposed in \citep{tscl}, which compared teacher approaches on a set of navigation tasks in Minecraft \citep{malmo}. While their work focuses on discrete sets of environments, we tackle the broader challenge of dealing with continuous parameter spaces that map to distributions over environments, and in which large parts of the parameter space may be unlearnable.


Another form of CL has already been studied for continuous sets of tasks \citep{goalgan}, however they considered goals (varying by their reward function), where we tackle the more complex setting of learning to behave in a continuous set of environments (varying by their state space). The GOAL-GAN algorithm also requires to set a reward range of "intermediate difficulty" to be able to label each goal in order to train the GAN, which is highly dependent on both the learner's skills and the considered continuous set of goals. Besides, as the notion of intermediate difficulty provides no guarantee of progress, this approach is susceptible to focusing on unlearnable goals for which the learner's competence stagnates in the intermediate difficulty range.


\section{The Continuous Teacher-Student Framework (CTS)}
\label{sec:framework}

 In this section, we formalize our Continuous Teacher-Student framework. It is inspired from earlier work in developmental robotics \citep{riac, strategicstudent} and intelligent tutoring systems \citep{zpdes}. The CTS framework is also close to earlier work on Teacher-Student approaches for discrete sets of tasks \citep{tscl}. In CTS however, teachers sample parameters mapping to \textit{distributions} of tasks from a continuous parameter space. In the remainder of this paper, we will refer to parameter sampling and task distribution sampling interchangeably, as one parameter directly maps to a task distribution.

\paragraph{Student} In CTS, learning agents, called students, are confronted with episodic Partially Observable Markov Decision Processes (POMDP) tasks $\tau$. For each interaction step in $\tau$, a student collects an observation $o \in \mathcal{O}_\tau$, performs an action $a \in \mathcal{A_\tau}$, and receives a corresponding reward $r \in \mathbb{R_\tau}$. Upon task termination, an episodic reward $r_e=\sum_{t=0}^{T} r^{(t)}$ is computed, with $T$ the length of the episode.

\paragraph{Teacher} The teacher interacts with its student by selecting a new parameter $p \in \mathcal{P}$, mapped to a task distribution $\mathcal{T}(p) \in \mathcal{T}$, proposing $m$ tasks $\tau \sim \mathcal{T}(p)$ to its student and observing $r_p$, the average of the $m$ episodic rewards $r_e$. The new parameter-reward tuple is then added to an history database of interactions $\mathcal{H}$ that the teacher leverages to influence the parameter selection in order to maximize the student's final competence return $c_p=\mathit{f}(r_p)$ across the parameter space. Formally, the objective is
\begin{equation}\label{eq:1}
    max \int_{\mathcal{P}, t=K} \! w_p \cdot c_p^K\, \mathrm{d}p,
\end{equation}
with $K$ the predefined maximal number of teacher-student interactions and $w_p$, a factor weighting the relative importance of each task distribution in the optimization process, enabling to specify whether to focus on specific subregions of the parameter space (i.e. harder target tasks). As students are considered as black-box learners, the teacher solely relies on its database history $\mathcal{H}$ for parameter sampling and does not have access to information about its student's internal state, algorithm, or perceptual and motor capacities. 


\paragraph{Parameter space assumptions} The teacher does not know the evolution of difficulty across the parameter space and therefore assumes a non-linear, piece-wise smooth function. The parameter space may also be ill-defined. For example, there might be subregions  $\mathcal{U} \subset \mathcal{P}$ of the parameter space in which competence improvements on any parameter $u \in \mathcal{U}$ is not possible given the state transition functions $\mathcal{F}_\tau$ of tasks sampled in any $\mathcal{T}(u) \in \mathcal{T}$ (i.e tasks are either trivial or unfeasible). Additionally, given a parameter space $\mathcal{P} \in \mathbb{R}^d$, there might exist an equivalent parameter space $\mathcal{P'} \in \mathbb{R}^k$ with $k < d$, constructed with a subset of the $d$ dimensions of $\mathcal{P}$, meaning that there might be irrelevant or redundant dimensions in $\mathcal{P}$.


In the following sections, we will restrict our study to CTS setups in which teachers sample only one task per selected parameter vector (i.e $m=1$ and $r_p$ = $r_e$) and do not prioritize the learning of specific subspaces (i.e $w_p = 1,\forall p \in P$).


\section{Methods}
\label{sec:methods}


In this section, we will describe our absolute LP-based teacher algorithms, our reference teachers, and present the continuously parameterized BipedalWalker environments used to evaluate them.

\subsection{Absolute Learning-Progress-Based Teacher Algorithms}
\label{ssec:alpalgos}


Of central importance to this paper is the concept of learning progress, formulated as a theoretical hypothesis to account for intrinsically motivated learning in humans \citep{kaplan2007search}, and applied for efficient robot learning \citep{devrobbook, playgroundexp}. Inspired by some of this work \citep{baranes2013active, imgep}, we frame our two teacher approaches as a Multi-Armed Bandit setup in which arms are dynamically mapped to subspaces of the parameter space, and whose values are defined by an absolute average LP utility function. The objective is then to select subspaces on which to sample a distribution of tasks in order to maximize ALP. ALP gives a richer signal than (positive) LP as it enables the teacher to detect when a student is losing competence on a previously mastered parameter subspace (thus preventing catastrophic forgetting).


\paragraph{Robust Intelligent Adaptive Curiosity (RIAC)} RIAC \citep{riac} is a task sampling approach whose core idea is to split a given parameter space in hyperboxes (called regions) according to their absolute LP, defined as the difference of cumulative episodic reward between the newest and oldest tasks sampled in the region. Tasks are then sampled within regions selected proportionally to their ALP score. This approach can easily be translated to the problem of sampling distributions of tasks, as is the case in this work. To avoid a known tendency of RIAC to oversplit the space \citep{goalgan}, we added a few minor modifications to the original architecture to constrain the splitting process. Details can be found in appendix \ref{ann:impldetails}.

\paragraph{Absolute Learning Progress Gaussian Mixture Model (ALP-GMM)}
Another more principled way of sampling tasks according to LP measures is to rely on the well known Gaussian Mixture Model \citep{gmm} and Expectation-Maximization \citep{EMalgo} algorithms. This concept has already been successfully applied in the cognitive science field as a way to model intrinsic motivation in early vocal developments of infants \citep{moulinfriergmm}. In addition of testing for the first time their approach (referred to as Covar-GMM) on DRL students, we propose a variant based on an ALP measure capturing long-term progress variations that is well-suited for RL setups. See appendix \ref{ann:impldetails} for a description of their method.

The key concept of ALP-GMM is to fit a GMM on a dataset of previously sampled parameters concatenated to their respective ALP measure. Then, the Gaussian from which to sample a new parameter is chosen using an EXP4 bandit scheme \citep{auer2002nonstochastic} where each Gaussian is viewed as an arm, and ALP is its utility. This enables the teacher to bias the parameter sampling towards high-ALP subspaces. To get this per-parameter ALP value, we take inspiration from earlier work on developmental robotics \citep{imgep}: for each newly sampled parameter $p_{new}$ and associated episodic reward $r_{new}$, the closest (Euclidean distance) previously sampled parameter $p_{old}$ (with associated episodic reward $r_{old}$) is retrieved using a nearest neighbor algorithm (implemented with a KD-Tree \citep{kdtree}). We then have \begin{equation}
\label{eq:2}
    alp_{new} = |r_{new} - r_{old}|
\end{equation}

The GMM is fit periodically on a window $\mathcal{W}$ containing only the most recent parameter-ALP pairs (here the last $250$) to bound its time complexity and make it more sensitive to recent high-ALP subspaces. The number of Gaussians is adapted online by fitting multiple GMMs (here having from $2$ to $k_{max}=10$ Gaussians) and keeping the best one based on Akaike's Information Criterion \citep{aic}. Note that the nearest neighbor computation of per-parameter ALP uses a database that contains all previously sampled parameters and associated episodic rewards, which prevents any forgetting of long-term progress. In addition to its main task sampling strategy, ALP-GMM also samples random parameters to enable exploration (here $p_{rnd}=20\%$). See Algorithm \ref{algo:ALP-GMM} for pseudo-code and appendix \ref{ann:impldetails} for a schematic view of ALP-GMM.


\begin{algorithm}[H]
	\caption{~ Absolute Learning Progress Gaussian Mixture Model (ALP-GMM)}
	\label{algo:ALP-GMM}
	\begin{algorithmic}[1]
	
	\Require Student $\mathcal{S}$, parametric procedural environment generator $E$, bounded parameter space $\mathcal{P}$, probability of random sampling $p_{rnd}$, fitting rate $N$, max number of Gaussians $k_{max}$
	\vspace{0.2cm}
	\State Initialize parameter-ALP First-in-First-Out window $\mathcal{W}$, set max size to $N$
	\State Initialize parameter-reward history database $\mathcal{H}$
	\Loop~$N$ times \Comment Bootstrap phase
	    \State Sample random $p \in \mathcal{P}$, send $E(\tau \sim \mathcal{T}(p))$ to $\mathcal{S}$, observe episodic reward $r_p$
	    \State Compute ALP of $p$ based on $r_p$ and $\mathcal{H}$ (see equation \ref{eq:2})
	    \State Store $(p,r_p)$ pair in $\mathcal{H}$, store $(p,ALP_p)$ pair in $\mathcal{W}$
	\EndLoop
	\Loop \Comment Stop after $K$ inner loops
	\State Fit a set of GMM having 2 to $k_{max}$ kernels on $\mathcal{W}$
	\State Select the GMM with best Akaike Information Criterion
	\Loop~$N$ times
	    \State $p_{rnd} \%$ of the time, sample a random parameter $p \in \mathcal{P}$
	    \State Else, sample $p$ from a Gaussian chosen proportionally to its mean ALP value 
		\State Send $E(\tau \sim \mathcal{T}(p))$ to student $\mathcal{S}$ and observe episodic reward $r_p$
	    \State Compute ALP of $p$ based on $r_p$ and $\mathcal{H}$
	    \State Store $(p,r_p)$ pair in $\mathcal{H}$, store $(p,ALP_p)$ pair in $\mathcal{W}$
	\EndLoop
	\EndLoop
	\State \textbf{Return} $\mathcal{S}$
	
	\end{algorithmic}
\end{algorithm}

\subsection{Teacher References}
\paragraph{Random Task Curriculum (Random)} In Random, parameters are sampled randomly in the parameter space for each new episode. Although simplistic, similar approaches in previous work \citep{baranes2013active} proved to be competitive against more elaborate forms of CL.

\paragraph{Oracle} A hand-constructed approach, sampling random task distributions in a fixed-size sliding window on the parameter space. This window is initially set to the easiest area of the parameter space and is then slowly moved towards complex ones, with difficulty increments only happening if a minimum average performance is reached. Expert knowledge is used to find the dimensions of the window, the amplitude and direction of increments, and the average performance threshold. Pseudo-code is available in Appendix \ref{ann:impldetails}.

\subsection{Parameterized BipedalWalker Environments with Procedural Generation}
\label{ssec:PBW}

The BipedalWalker environment \citep{gym} offers a convenient test-bed for continuous control, allowing to easily build parametric variations of the original version \citep{agentdesignha, poet}. The learning agent, embodied in a bipedal walker, receives positive rewards for moving forward and penalties for torque usage and angular head movements. Agents are allowed $2000$ steps to reach the other side of the map. Episodes are aborted with a $-100$ reward penalty if the walker's head touches an obstacle.

To study the ability of our teachers to guide DRL students, we design two continuously parameterized BipedalWalker environments enabling the procedural generation of walking tracks:
\begin{itemize}
    \item \textbf{Stump Tracks} A $2$D parametric environment producing tracks paved with stumps varying by their height and spacing. Given a parameter vector $[\mu_h, \Delta_s]$, a track is constructed by generating stumps spaced by $\Delta_s$ and whose heights are defined by independent samples in a normal distribution $\mathcal{N}(\mu_h, 0.1)$.
    
    \item \textbf{Hexagon Tracks} A more challenging $12$D parametric BipedalWalker environment. Given $10$ offset values $\mu_o$, each track is constructed by generating hexagons having their default vertices' positions perturbed by strictly positive independent samples in $\mathcal{N}(\mu_o, 0.1)$. The remaining $2$ parameters are distractors defining the color of each hexagon. This environment is challenging as there are no subspaces generating trivial tracks with 0-height obstacles (as offsets to the default hexagon shape are positive). This parameter space also has non-linear difficulty gradients as each vertices have different impacts on difficulty when modified.
\end{itemize}

All of the experiments done in these environments were performed using OpenAI's implementation  of Soft-Actor Critic \citep{sac} as the single student algorithm. To test our teachers' robustness to students with varying abilities, we use $3$ different walker morphologies (see Figure \ref{students}). Additional details on these two environments along with track examples are available in Appendix \ref{app:pbw}.

\begin{figure*}[htb!]
\centering
\subfloat[Stump Tracks walkers]{\includegraphics[height= 1.5cm]{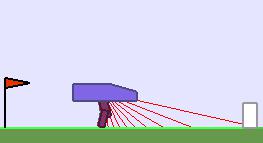}
                                \includegraphics[height= 1.5cm]{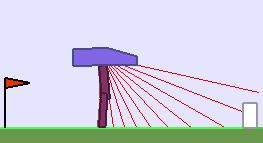}
                                \includegraphics[height=1.5cm]{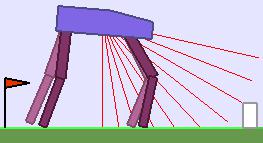}}
\hspace{1.5cm}
\subfloat[Hexagon Tracks walker]{\includegraphics[height= 1.5cm]{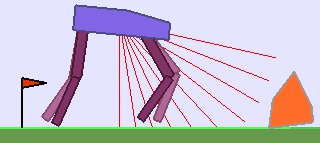}}
\caption{\footnotesize{\textbf{Multiple students and environments to benchmark teachers.} \textbf{(a):} In addition to the default bipedal walker morphology (middle agent), we designed a bipedal walker with 50\% shorter legs (left) and a bigger quadrupedal walker (right). \textbf{(b):} The quadrupedal walker is also used in Hexagon Tracks.}}
\label{students}
\end{figure*}


\section{Experimental Results}
\label{sec:results}

\paragraph{Performance metric} To assess the performance of all of our approaches on our BipedalWalker environments, we define a binary competence return measure stating whether a given track distribution is mastered or not, depending on the student's episodic reward $r_p$. We set the reward threshold to $230$, which was used in \citep{poet} to ensure "reasonably efficient" walking gates for default bipedal walkers trained on environments similar to ours. Note that this reward threshold is only used for evaluation purposes and in the Oracle condition. Performance is then evaluated periodically by sampling a single track in each track distribution of a fixed evaluation set of $50$ distributions sampled uniformly in the parameter space. We then simply measure the percentage of mastered tracks. During evaluation, learning in DRL agents is turned off. 

Through our experiments we answer three questions about ALP-GMM, Covar-GMM and RIAC:
\begin{itemize}
    \item Are ALP-GMM, Covar-GMM and RIAC able to optimize their students' performance better than random approaches and teachers exploiting environment knowledge?
    \item  How does their performance scale when the proportion of unfeasible tasks increases? 
    \item Are they able to scale to high-dimensional sampling spaces with irrelevant dimensions?  
\end{itemize}

\subsection{How do ALP-GMM, Covar-GMM and RIAC compare to reference teachers?}
\label{sec:results:1}

\begin{figure*}[htb!]
\centering
\subfloat[After 500 episodes]{\includegraphics[height=4.5cm]{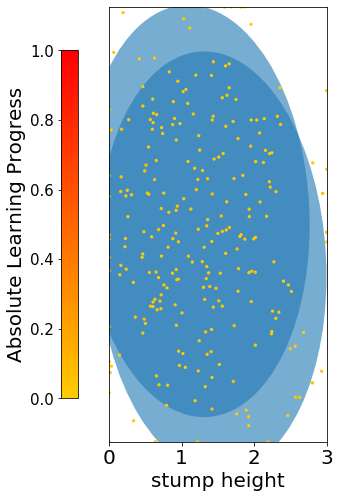}}
\subfloat[1500 eps.]{\includegraphics[height=4.5cm]{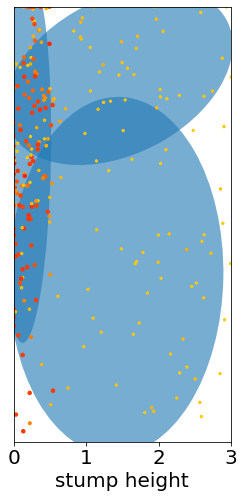}}
\subfloat[15000 eps.]{\includegraphics[height=4.5cm]{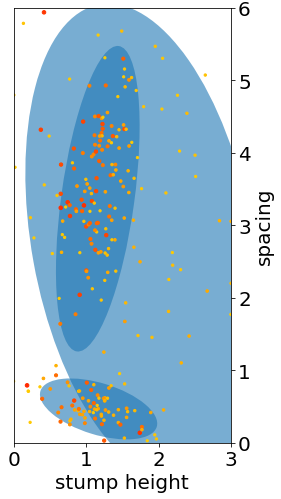}}
\subfloat[Mastered tracks (20M steps)]{\includegraphics[width=4.03cm]{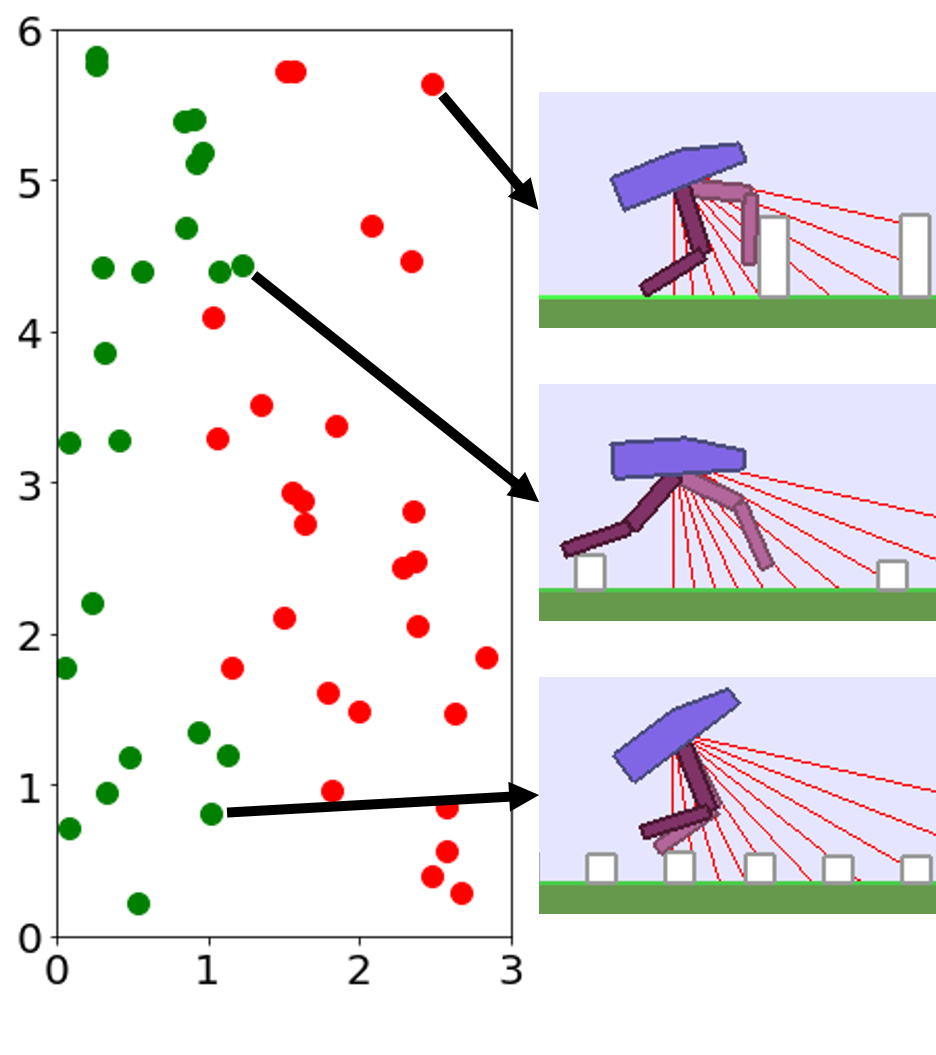}}
\caption{\footnotesize{\textbf{Example of an ALP-GMM teacher paired with a Soft Actor-Critic student on Stump Tracks.} Figures (a)-(c) show the evolution of ALP-GMM parameter sampling in a representative run. Each dot represents a sampled track distribution and is colored according to its Absolute Learning Progress value. After initial progress on the leftmost part of the space, as in (b), most ALP-GMM runs end up improving on track distributions with $1$ to $1.8$ stump height, with the highest ones usually paired with spacing above $2.5$ or below $1$, indicating that tracks with large or very low spacing are easier than those in $[1,2.5]$. Figure (d) shows for the same run which track distributions of the test set are mastered (i.e $r_t>230$, shown by green dots) after $17$k episodes.}}
\label{gmm_vizu}
\end{figure*}

\paragraph{Emergence of curriculum} 
Figure \ref{gmm_vizu} provides a visualization of the sampling trajectory observed in a representative ALP-GMM run for a default walker. Each plot shows the location of each Gaussian of the current mixture along with the $250$ track distributions subsequently sampled. At first (a), the walker does not manage to make any significant progress. After $1500$ episodes (b) the student starts making progress on the leftmost part of the parameter space, especially for track distributions with a spacing higher than $1.5$, which leads ALP-GMM to focus its sampling in that direction. After $15$k episodes (c) ALP-GMM has shifted its sampling strategy to more complex regions. The analysis of a typical RIAC run is detailed in Appendix \ref{ann:expdetails} (fig. \ref{riac_vizu}).

\begin{figure*}[htb!]
\centering
\subfloat[Short agents]{\includegraphics[width=4.65cm]{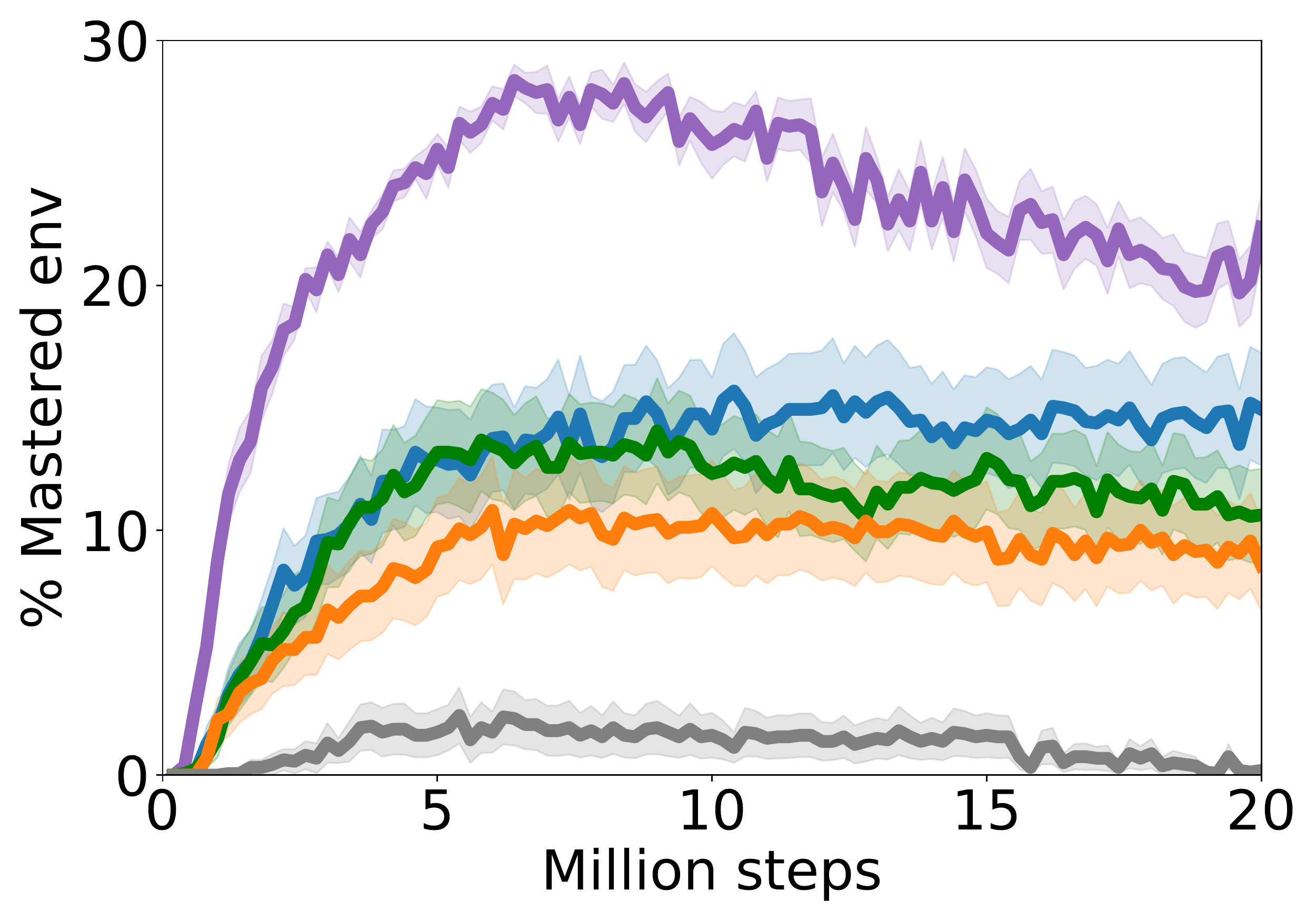}}
\subfloat[Default agents]{\includegraphics[width=4.65cm]{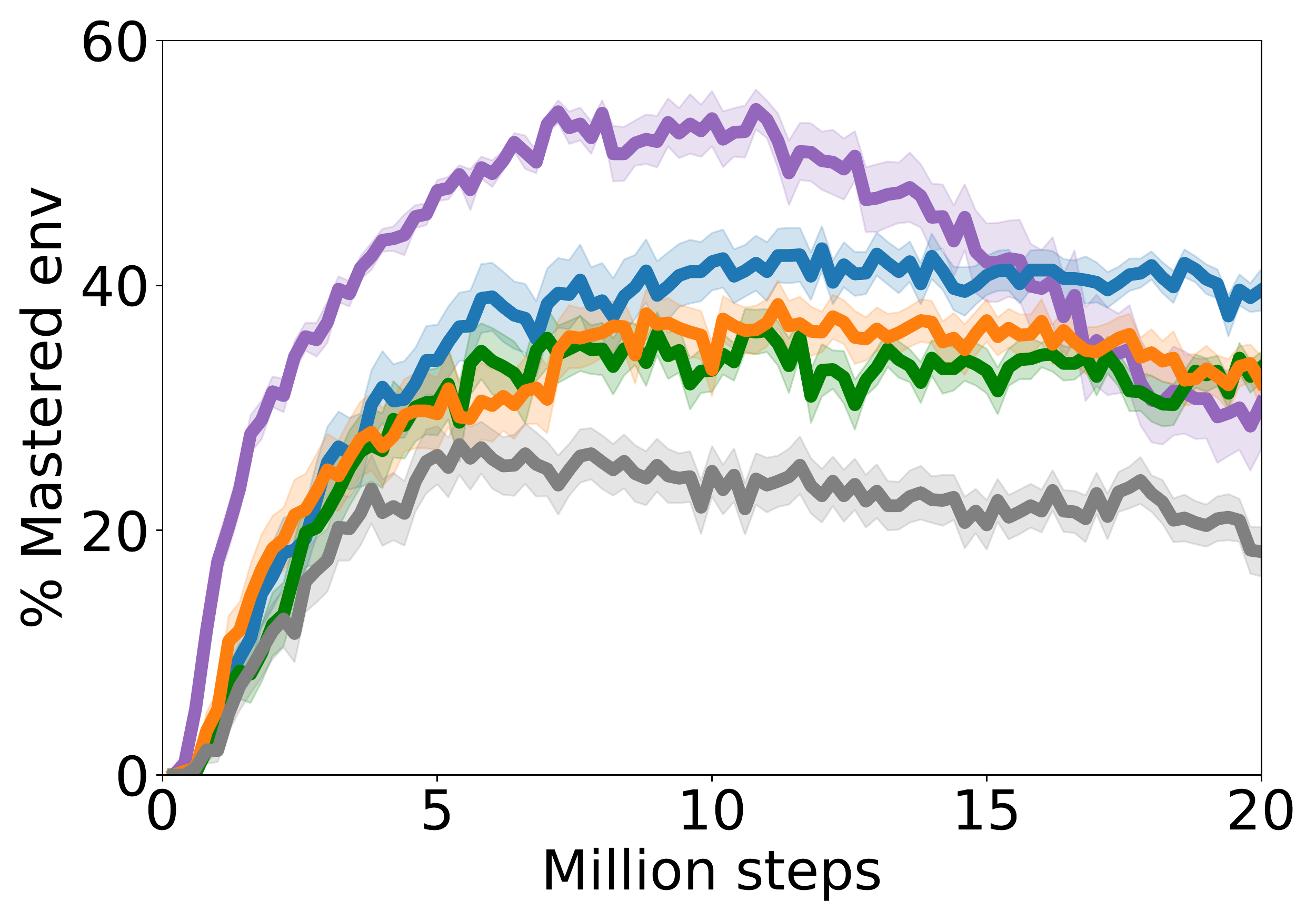}}
\subfloat[Quadrupedal agents]{\includegraphics[width=4.65cm]{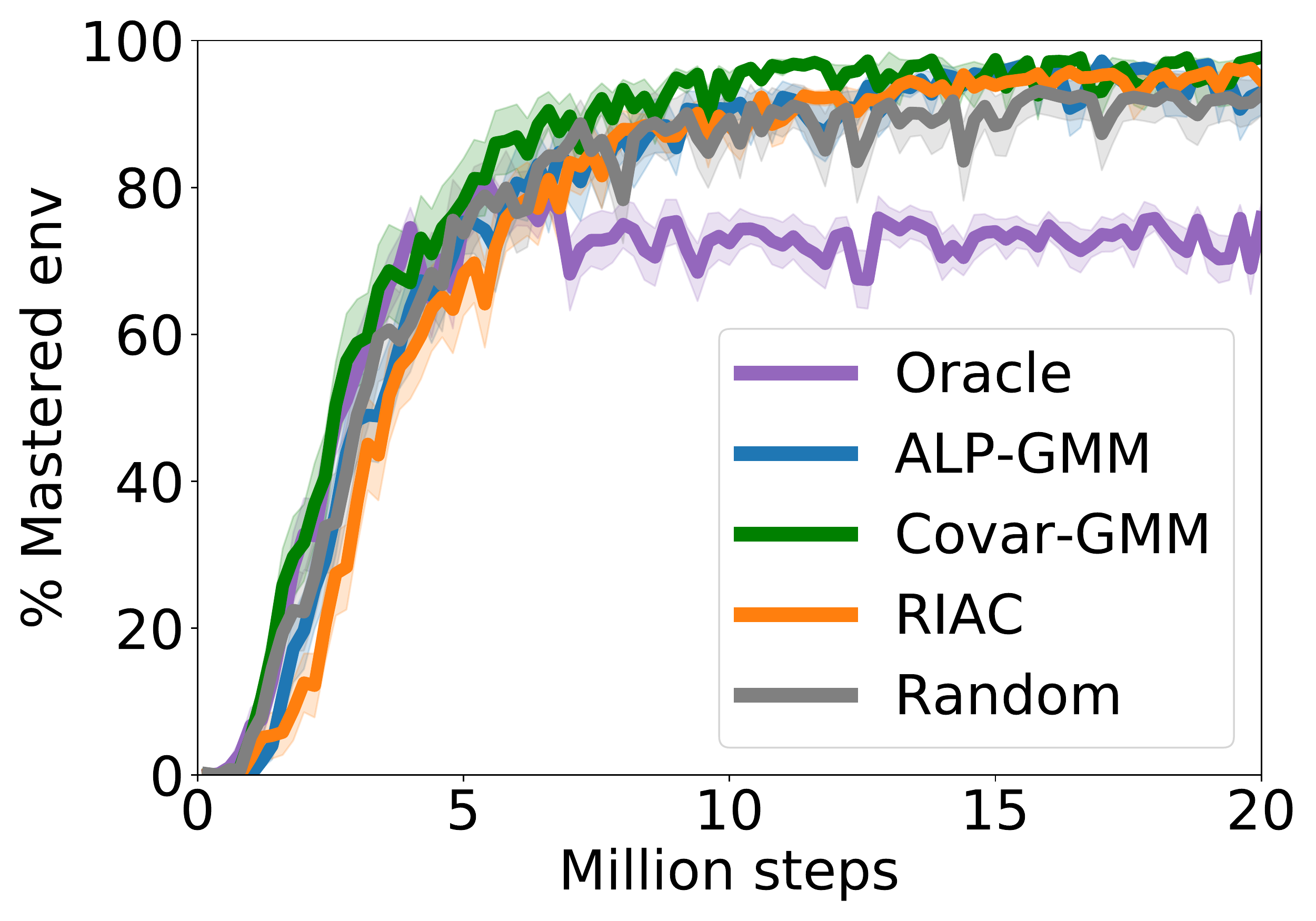}}
\caption{\footnotesize{\textbf{Evolution of mastered track distributions for Teacher-Student approaches in Stump Tracks.} The mean performance (32 seeded runs) is plotted with shaded areas representing the standard error of the mean.}}
\label{all_perfs_simple}
\end{figure*}

\paragraph{Performance comparison} Figure \ref{all_perfs_simple} shows learning curves for each condition paired with short, default and quadrupedal walkers. First of all, for short agents (a), one can see that Oracle is the best performing algorithm, mastering more than $20$\% of the test set after $20$ Million steps. This is an expected result as Oracle knows where to sample simple track distributions, which is crucial when most of the parameter space is unfeasible, as is the case with short agents. ALP-GMM is the LP-based teacher with highest final mean performance, reaching $14.9$\% against $10.6$\% for Covar-GMM and $8.6$\% for RIAC. This performance advantage for ALP-GMM is statistically significant when compared to RIAC (Welch's t-test at $20$M steps: $p<0.04$), however there is no statistically significant difference with Covar-GMM ($p=0.16$). All LP-based teachers are significantly superior to Random ($p<0.001$).


Regarding default bipedal walkers (b), our hand-made curriculum (Oracle) performs better than other approaches for the first $10$ Million steps and then rapidly decreases to end up with a performance comparable to RIAC and Covar-GMM. All LP-based conditions end up with a final mean performance statistically superior to Random ($p<10^{-4}$). ALP-GMM is the highest performing algorithm, significantly superior to Oracle ($p<0.04$), RIAC ($p<0.01$) and Covar-GMM ($p<0.01$).

For quadrupedal walkers (c), Random, ALP-GMM, Covar-GMM and RIAC agents quickly learn to master nearly $100$\% of the test set, without significant differences apart from Covar-GMM being superior to RIAC ($p<0.01$). This indicates that, for this agent type, the parameter space of Stump Tracks is simple enough that trying random tracks for each new episode is a sufficient curriculum learning strategy. Oracle teachers perform significantly worse than any other method ($p<10^{-5}$).

Through this analysis we answered our first experimental question by showing how ALP-GMM, Covar-GMM and RIAC, without strong assumptions on the environment, managed to scaffold the learning of multiple students better than Random. Interestingly, ALP-GMM outperformed Oracle with default agents, and RIAC, Covar-GMM and ALP-GMM surpassed Oracle with the quadrupedal agent, despite its advantageous use of domain knowledge. This indicates that training only on track distributions sampled from a sliding window that end up on the most difficult parameter subspace leads to forgetting of simpler task distributions. Our approaches avoid this issue through efficient tracking of their students' learning progress.

\subsection{How does our approaches scale when the amount of unfeasible tasks increases? }



A crucial requirement when designing all-purpose teacher algorithms is to ensure their ability to deal with parameter spaces that are ill-defined w.r.t to the considered student. To study this property we performed additional experiments on Stump Tracks where we gradually increased the stump height dimension range, which increases the amount of unfeasible tracks.

Results are summarized in Table \ref{table_stump_height}. To assess whether a condition is robust to increasing unfeasibility, one can look at the p-value of the Welch's t-test performed on the final performance measure between the condition run on the original parameter space and the same condition run on a wider space. High \mbox{p-value} indicates that there is not enough evidence to reject the null hypothesis of no difference, which can be interpreted as being robust to parameter spaces containing more unfeasible tasks. Using this metric, it is clear that ALP-GMM is the most robust condition among the presented LP-based teachers, with a p-value of $0.71$ when increasing the stump height range from $[0,3]$ to $[0,4]$ compared to $p=0.02$ for RIAC and $p=0.05$ for Covar-GMM. When going from $[0,3]$ to $[0,5]$, ALP-GMM is the only LP-based teacher able to maintain most of its performance ($p=0.12$). Although Random also seems to show robustness to increasingly unfeasible parameter spaces ($p=0.78$ when going from $[0,3]$ to $[0,4]$ and $p=0.05$ from $[0,3]$ to $[0,5]$), it is most likely due to its stagnation in low performances. Compared to all other approaches, ALP-GMM remains the highest performing condition in both parameter space variations ($p<0.02$).

\begin{table}[H]
\centering
\setlength\tabcolsep{4pt}
\begin{tabular}{|l|c|c|c|}
  \hline
  Cond. \textbackslash~Stump height  & $[0,3]$ & $[0,4]$ & $[0,5]$ \\
  \hline
  ALP-GMM & $\mathbf{39.6}\pm9.6$ & $\mathbf{38.6}\pm11.6$ ($p=0.71$)& $\mathbf{34.3}\pm15.8$ ($p=0.12$) \\
  \hline
  Covar-GMM & $33.3\pm7.1$ & $27.6\pm14.0$ ($p=0.05$)& $24.1\pm14.7$ ($p=0.01$) \\
  \hline
  RIAC & $32.1\pm12.2$ & $23.2\pm17.2$($p=0.02$) & $20.5\pm15.4$ ($p=0.002$)\\
  \hline
  Random & $18.2\pm11.5$ & $17.4\pm 11.8$ ($p=0.78$)& $12.6\pm11.0$ ($p=0.05$)\\
    \hline
\end{tabular}
\vspace{0.3cm}
\caption{\footnotesize{\textbf{Impact of increasing the proportion of unfeasible tasks.} The average performance with standard deviation (after 20 Million steps) on the original Stump Tracks' test-set is reported (32 seeds per condition).The additional p-values inform whether conditions run in the original Stump Tracks ($[0,3]$) are significantly better than when run on variations with higher maximal stump height.}}
\label{table_stump_height}
\end{table}

These additional experiments on Stump Tracks showed that our LP-based teachers are able to partially maintain the performance of their students in parameter spaces with higher proportions of unfeasible tasks, with a significant advantage for ALP-GMM.

\subsection{Are our approaches able to scale to ill-defined high-dimensional task spaces?}

\begin{figure*}[htb!]
\centering
\subfloat{\includegraphics[width=6.0cm]{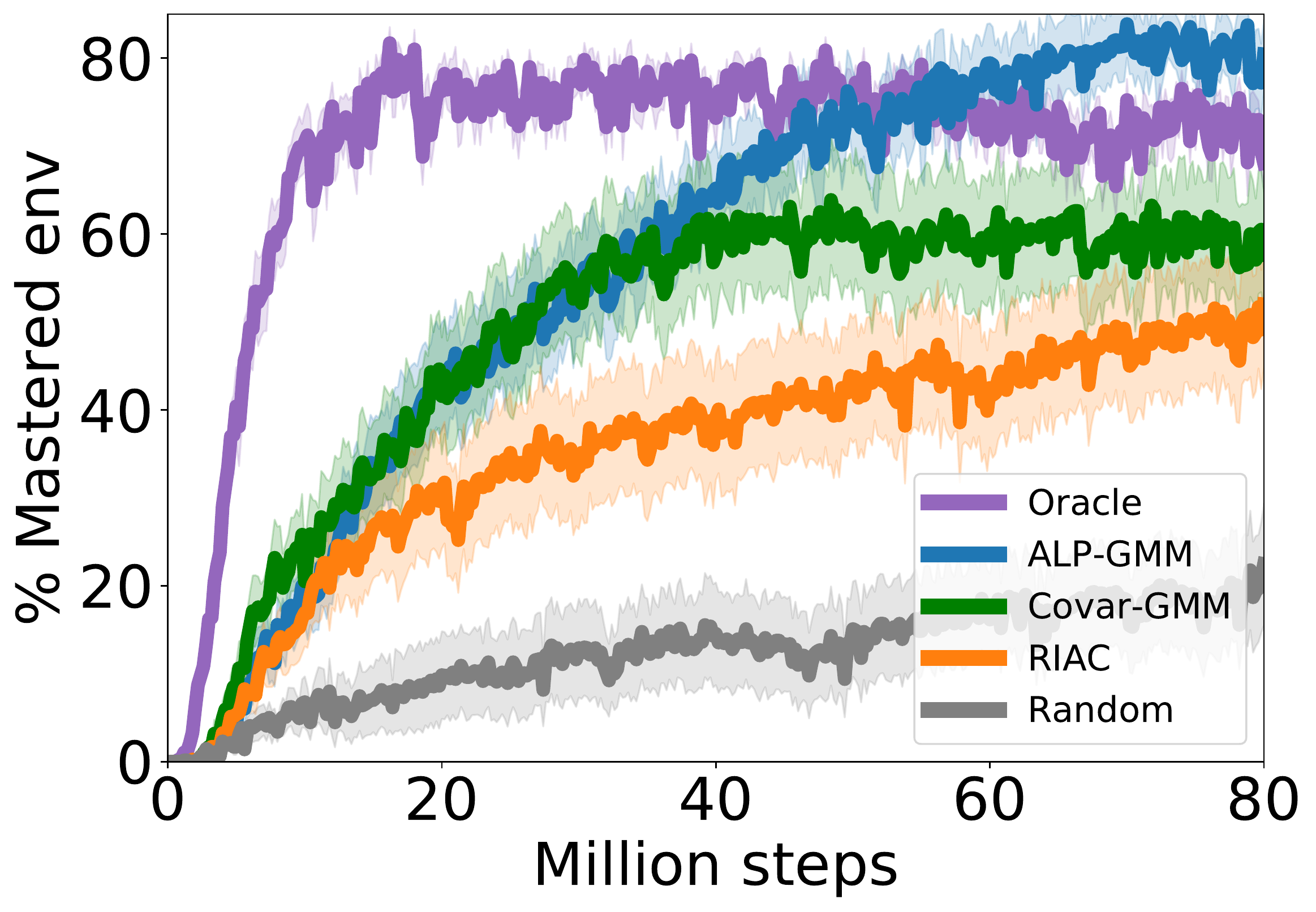}}
\subfloat{\includegraphics[width=6.0cm]{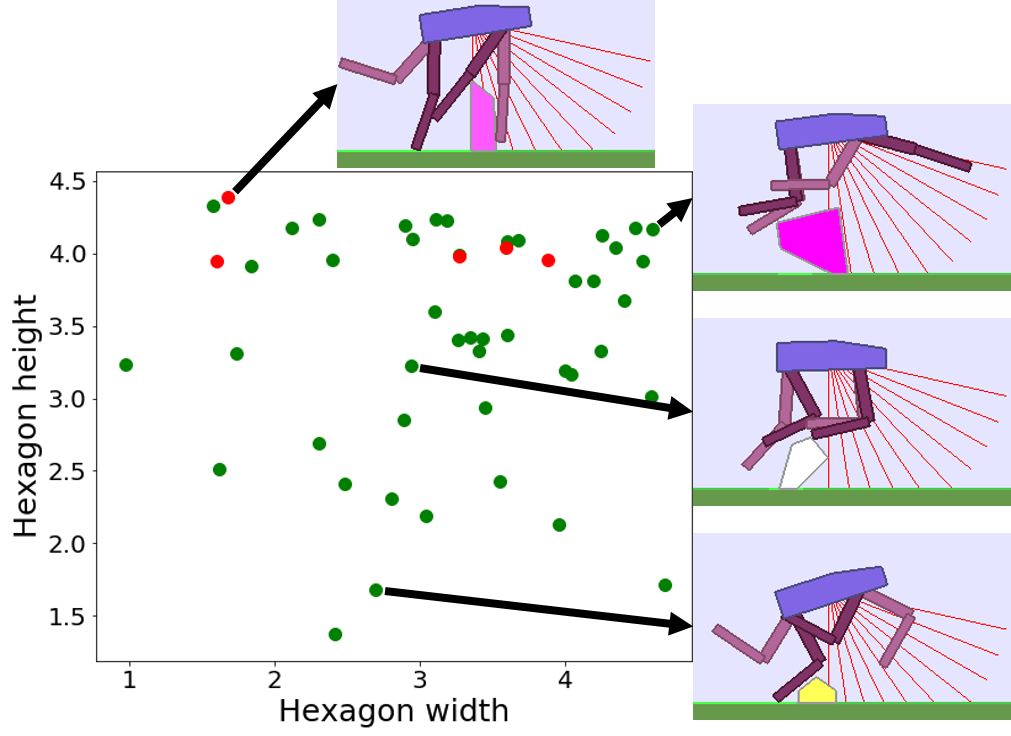}}
\caption{\footnotesize{\textbf{Teacher-Student approaches in Hexagon Tracks.} \textbf{Left:} Evolution of mastered tracks for Teacher-Student approaches in Hexagon Tracks. 32 seeded runs (25 for Random) of 80 Millions steps where performed for each condition. The mean performance is plotted with shaded areas representing the standard error of the mean. \textbf{Right:} A visualization of which track distributions of the test-set are mastered (i.e $r_t>230$, shown by green dots) by an ALP-GMM run after $80$ million steps.}}
\label{all_perfs_complex}
\end{figure*}

To assess whether ALP-GMM, Covar-GMM and RIAC are able to scale to parameter spaces of higher dimensionality containing irrelevant dimensions, and whose difficulty gradients are non-linear, we performed experiments with quadrupedal walkers on Hexagon Tracks, our 12-dimensional parametric BipedalWalker environment. Results are shown in Figure \ref{all_perfs_complex}. In the first $20$ Millions steps, one can see that Oracle has a large performance advantage compared to LP-based teachers, which is mainly due to its knowledge of initial progress niches. However, by the end of training, ALP-GMM significantly outperforms Oracle ($p<0.02$), reaching an average final performance of $80$\% against $68\%$ for Oracle. Compared to Covar-GMM and RIAC, the final performance of ALP-GMM is also significantly superior ($p<0.01$ and $p<0.005$, respectively) while being more robust and having less variance (see appendix \ref{ann:expdetailshexa}). All LP-based approaches are significantly better than Random ($p<0.01$).




Experiments on the Hexagon Tracks showed that ALP-GMM is the most suitable condition for complex high-dimensional environments containing irrelevant dimensions, non-linear parameter spaces and large proportions of initially unfeasible tasks.


\paragraph{Complementary experiments} To better grasp the general properties of our teacher algorithms, additional abstract experiments without DRL students were also performed for parameter spaces with increasing number of dimensions (relevant and irrelevant) and increasing ratio of initially unfeasible subspaces, showing that GMM-based approaches performed best (see Appendix \ref{app:gridworld}).



\section{Discussion and Conclusion}

This work demonstrated that LP-based teacher algorithms could successfully guide DRL agents to learn in difficult continuously parameterized environments with irrelevant dimensions and large proportions of unfeasible tasks. With no prior knowledge of its student's abilities and only loose boundaries on the task space, ALP-GMM, our proposed teacher, consistently outperformed random heuristics and occasionally even expert-designed curricula.

ALP-GMM, which is conceptually simple and has very few crucial hyperparameters, opens-up exciting perspectives inside and outside DRL for curriculum learning problems. Within DRL, it could be applied to previous work on autonomous goal exploration through incremental building of goal spaces \cite{Finot2019}. In this case several ALP-GMM instances could scaffold the learning agent in each of its autonomously discovered goal spaces. Another domain of applicability is assisted education, for which current state of the art relies heavily on expert knowledge \citep{zpdes} and is mostly applied to discrete task sets.

\clearpage
\acknowledgments{This work was supported by Microsoft Research through its PhD Scholarship Programme. Experiments were carried out using the PlaFRIM experimental testbed, supported by Inria, CNRS (LABRI and IMB), Université de Bordeaux, Bordeaux INP and Conseil Régional d'Aquitaine (see https://www.plafrim.fr/) and the Curta platform (see https://redmine.mcia.fr/).}


\bibliography{example}  

\newpage
\appendix

\section{Experiments on an n-dimensional toy parameter space}
\label{app:gridworld}

\paragraph{The n-dimensional toy space}
An $n$-dimensional toy parameter space $\mathcal{P} \in [0,1]^n$ was implemented to simulate a student learning process, enabling the study of our teachers in a controlled deterministic environment without DRL agents. A parameter $p$ directly maps to an episodic reward $r_p$ depending on the history of previously sampled parameters. The parameter space is divided in hypercubes and enforces the following rules:
\begin{itemize}
    \item Sampling a parameter in an "unlocked" hypercube results in a positive reward ranging from $1$ to $100$ depending on the amount of already sampled parameters in the hypercube: if $10$ parameters were sampled in it, the next one will yield a reward of $11$. Sampling a parameter located in a "locked" hypercube does not yield any reward.
    \item At first, all hypercubes are "locked" except for one, located in a corner.
    \item Sampling $75$ parameters in an unlocked hypercube unlocks its neighboring hypercubes.
\end{itemize}

\begin{figure*}[htb!]
\centering
\subfloat{\includegraphics[width=0.5\textwidth]{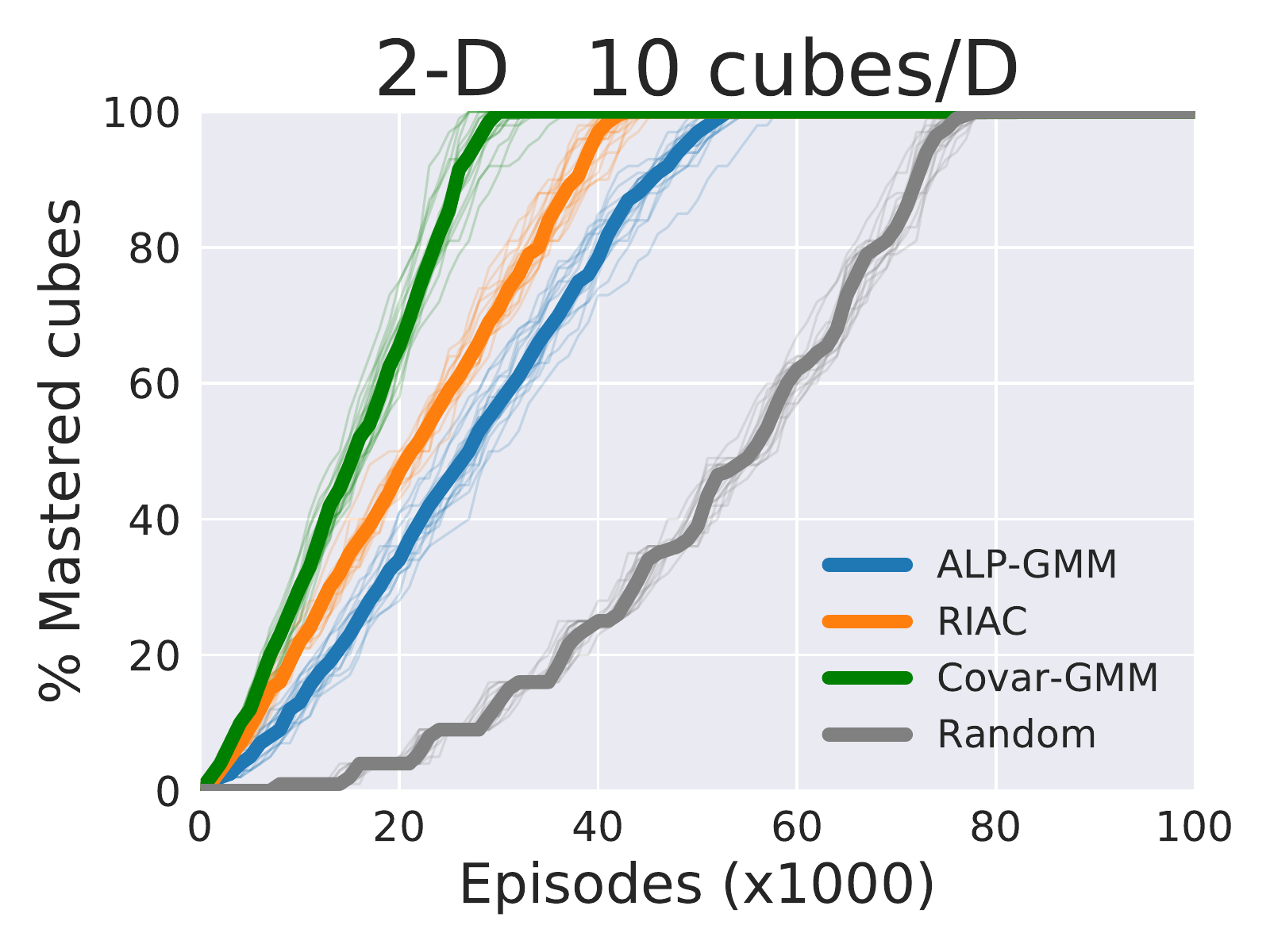}}

\subfloat{\includegraphics[width=4.65cm]{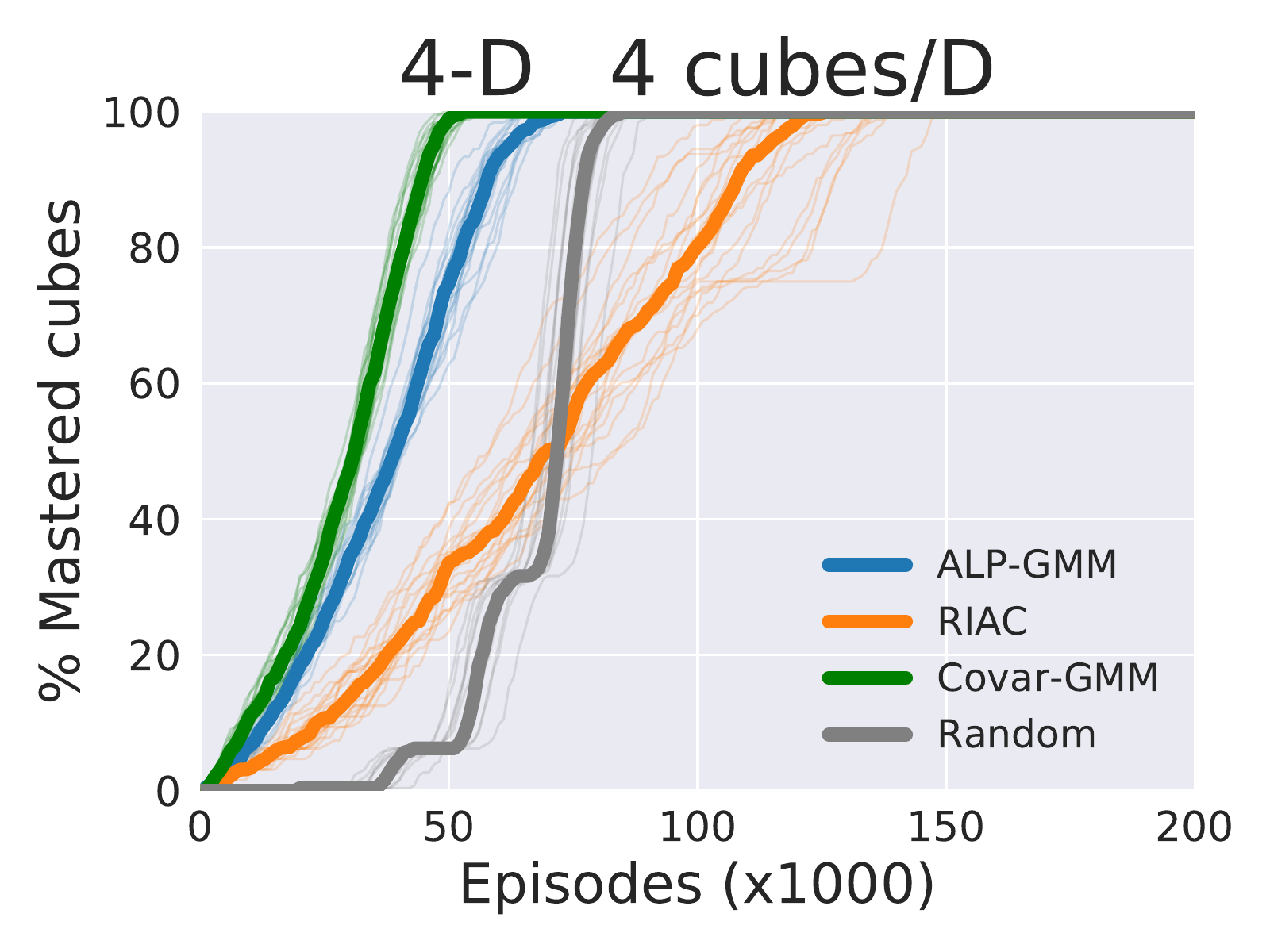}}
\subfloat{\includegraphics[width=4.65cm]{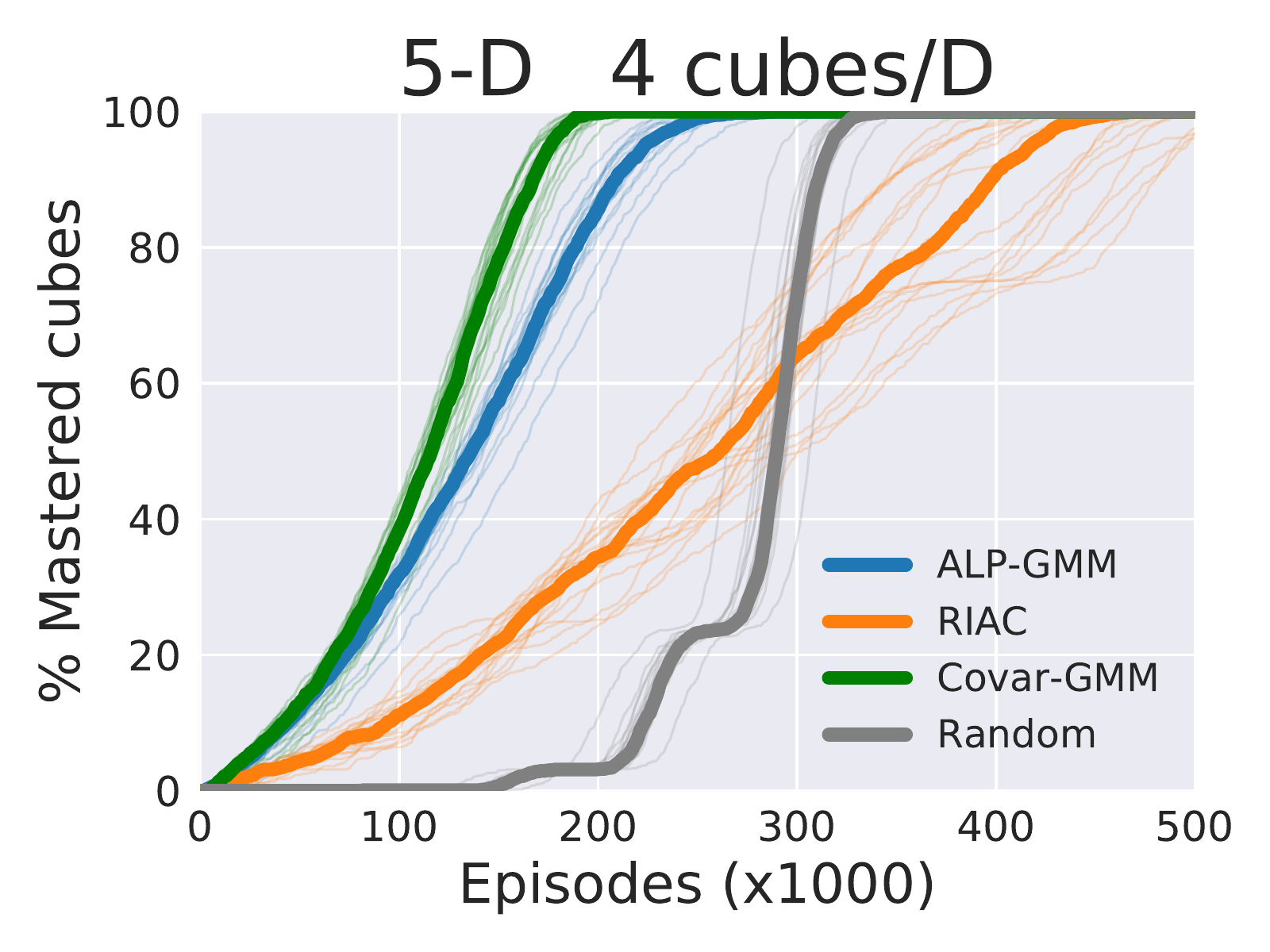}}
\subfloat{\includegraphics[width=4.65cm]{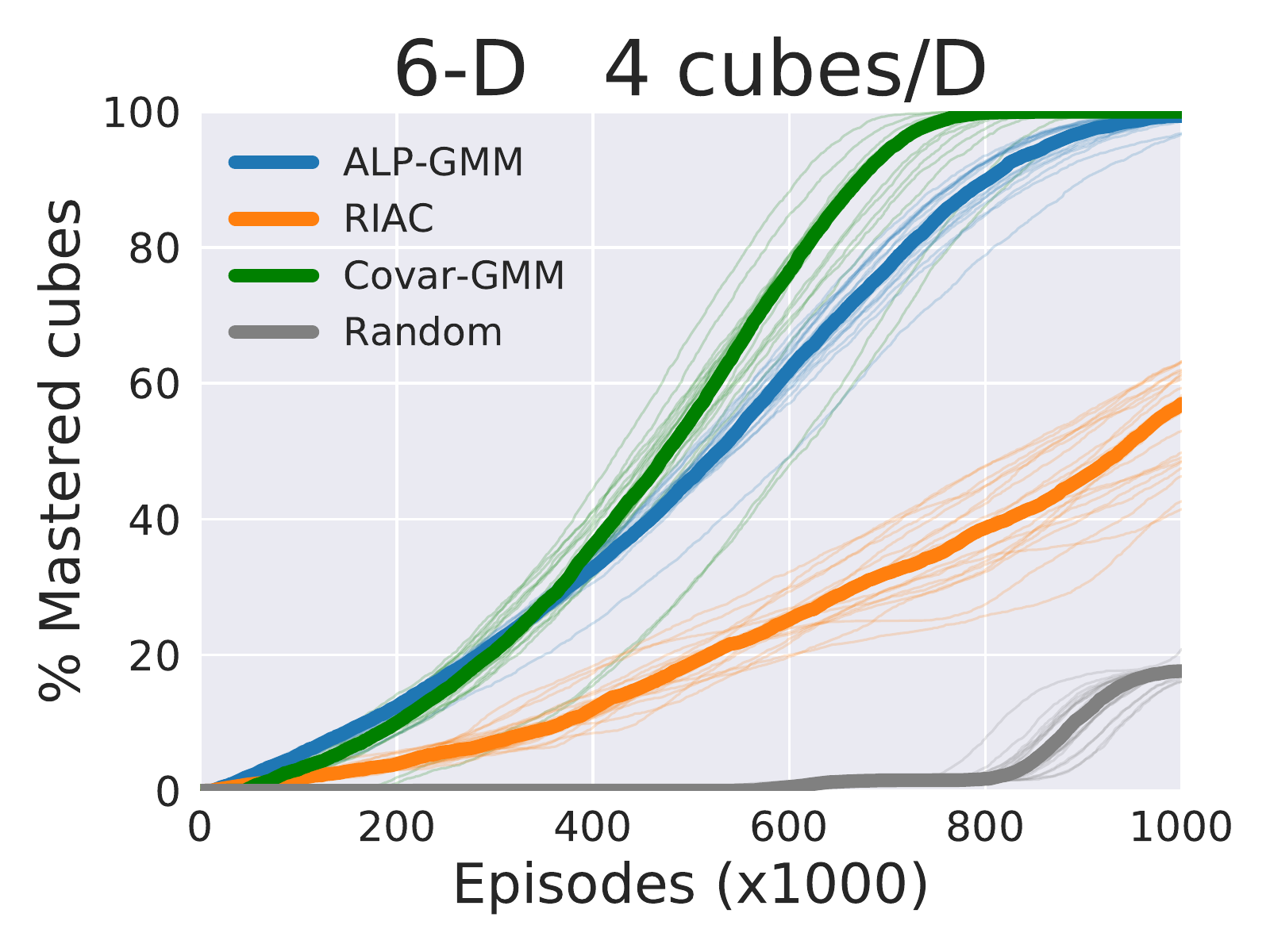}}

\subfloat{\includegraphics[width=4.65cm]{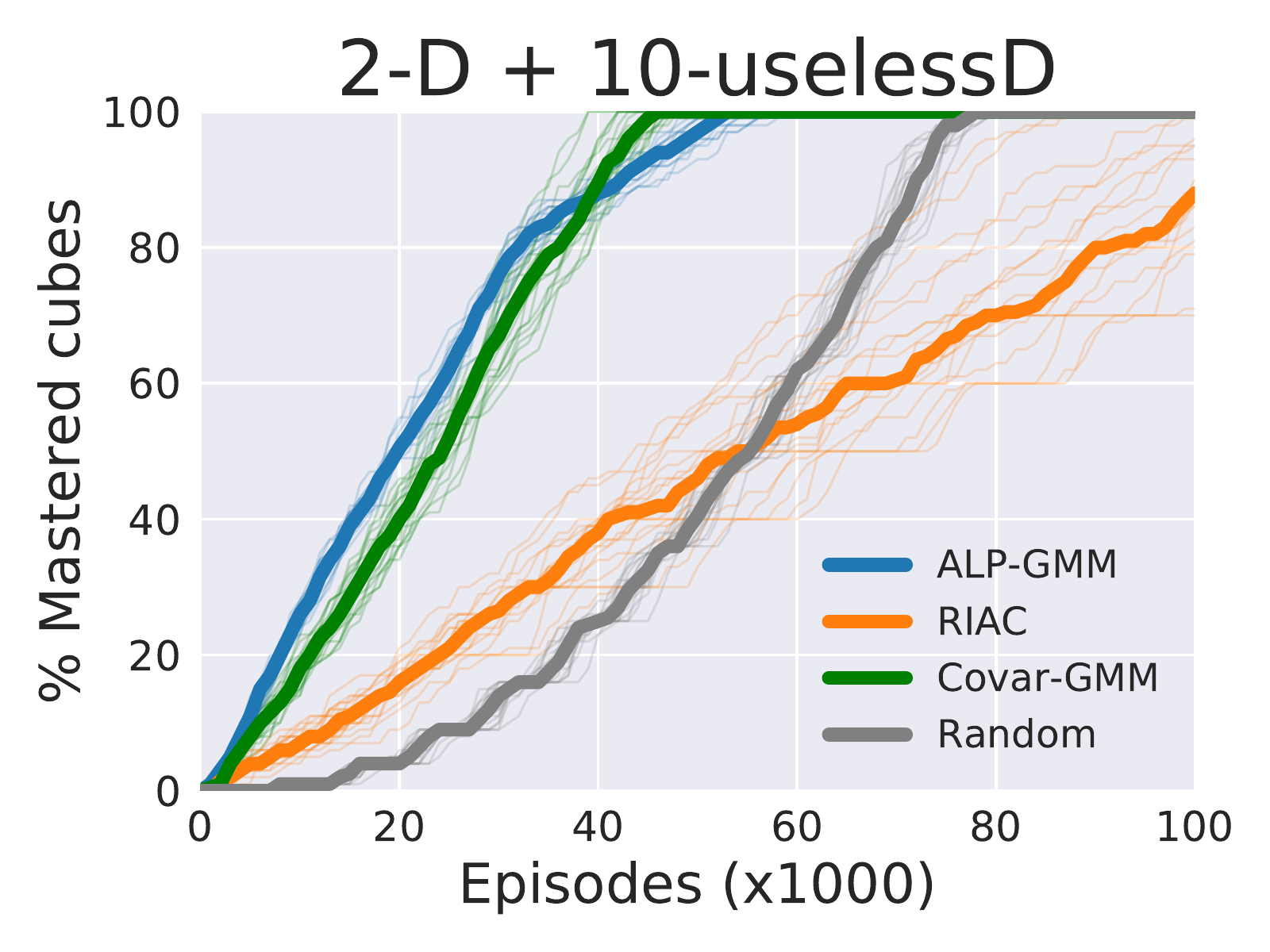}}
\subfloat{\includegraphics[width=4.65cm]{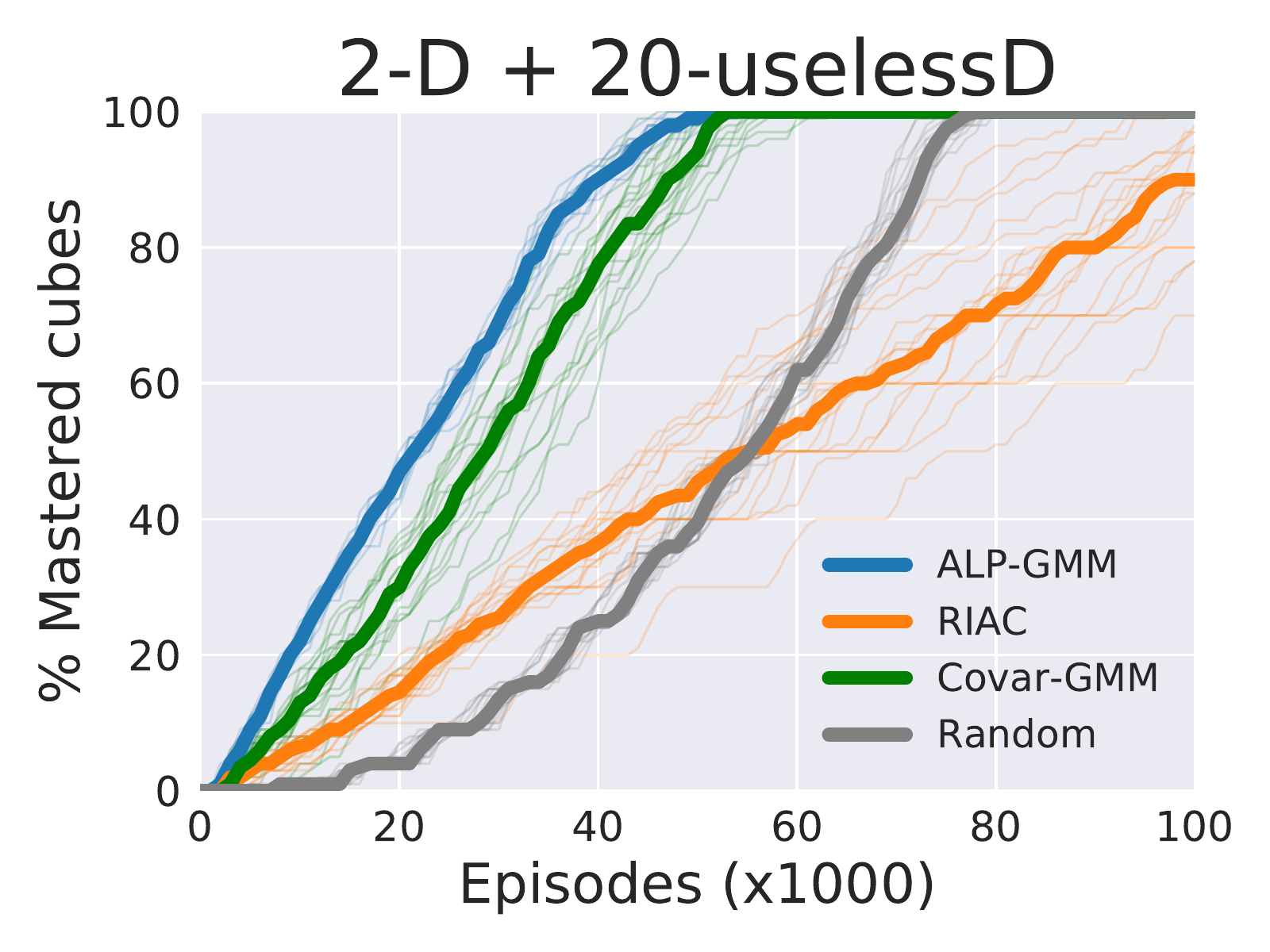}}
\subfloat{\includegraphics[width=4.65cm]{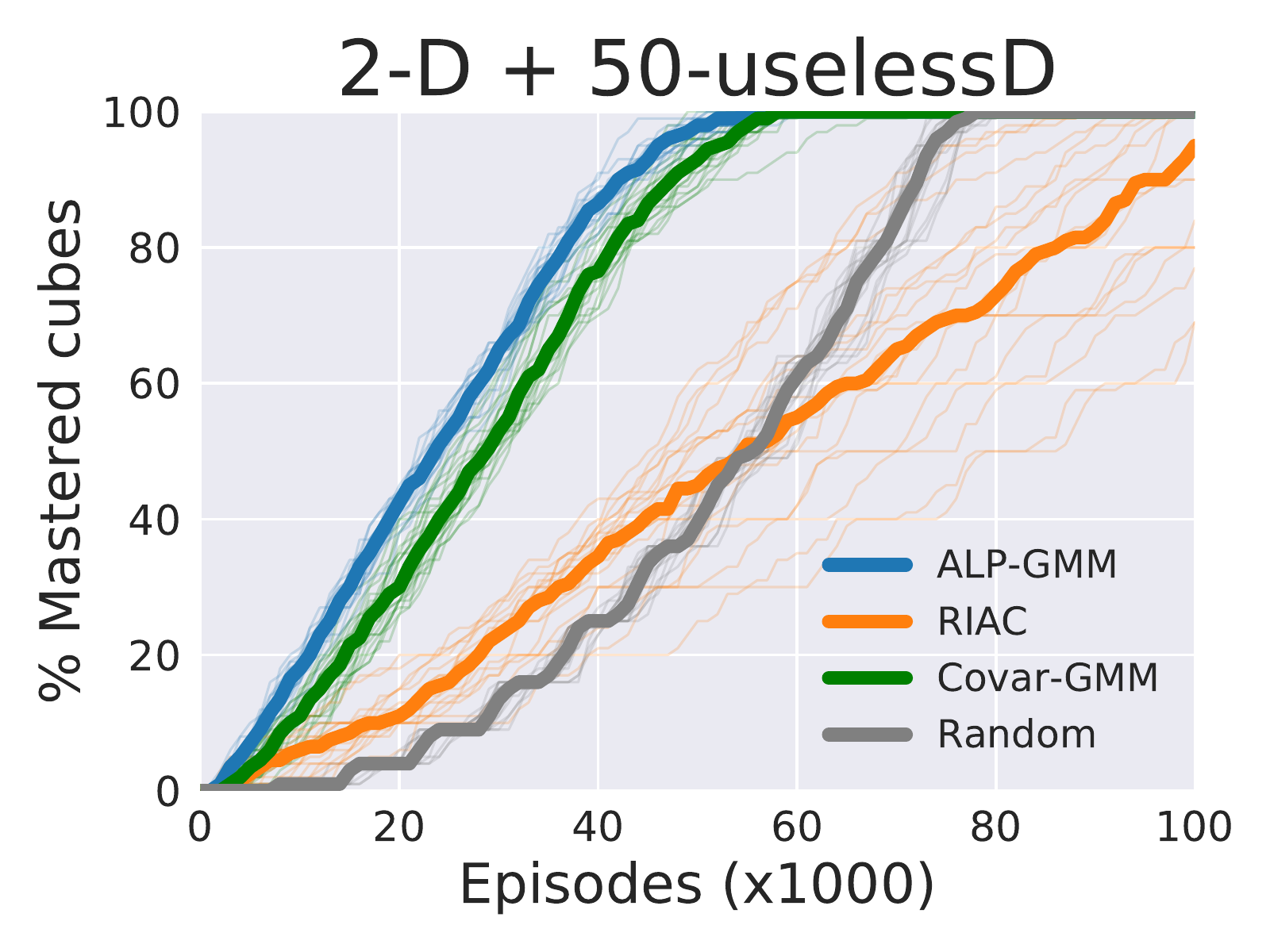}}

\subfloat{\includegraphics[width=4.65cm]{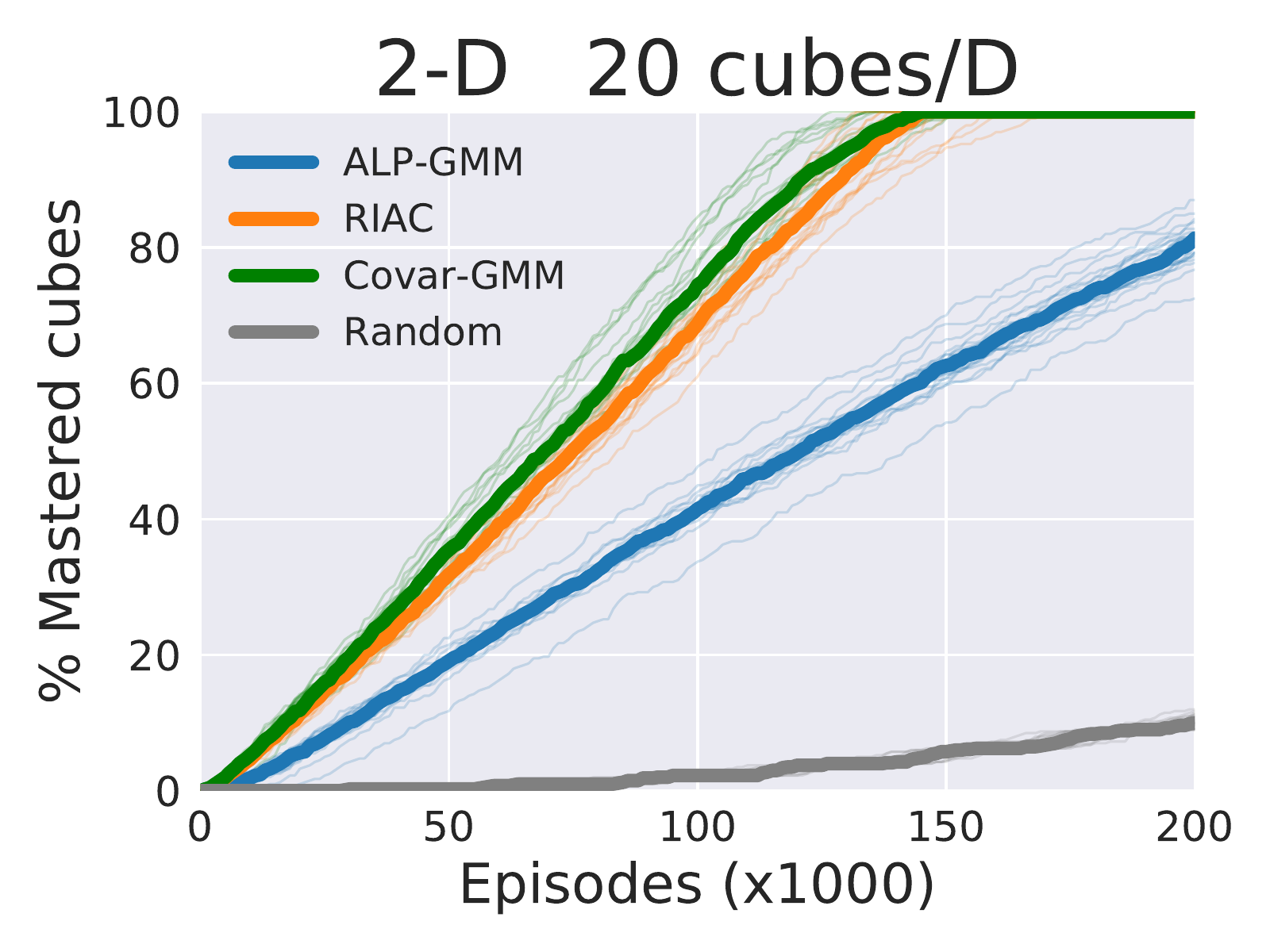}}
\subfloat{\includegraphics[width=4.65cm]{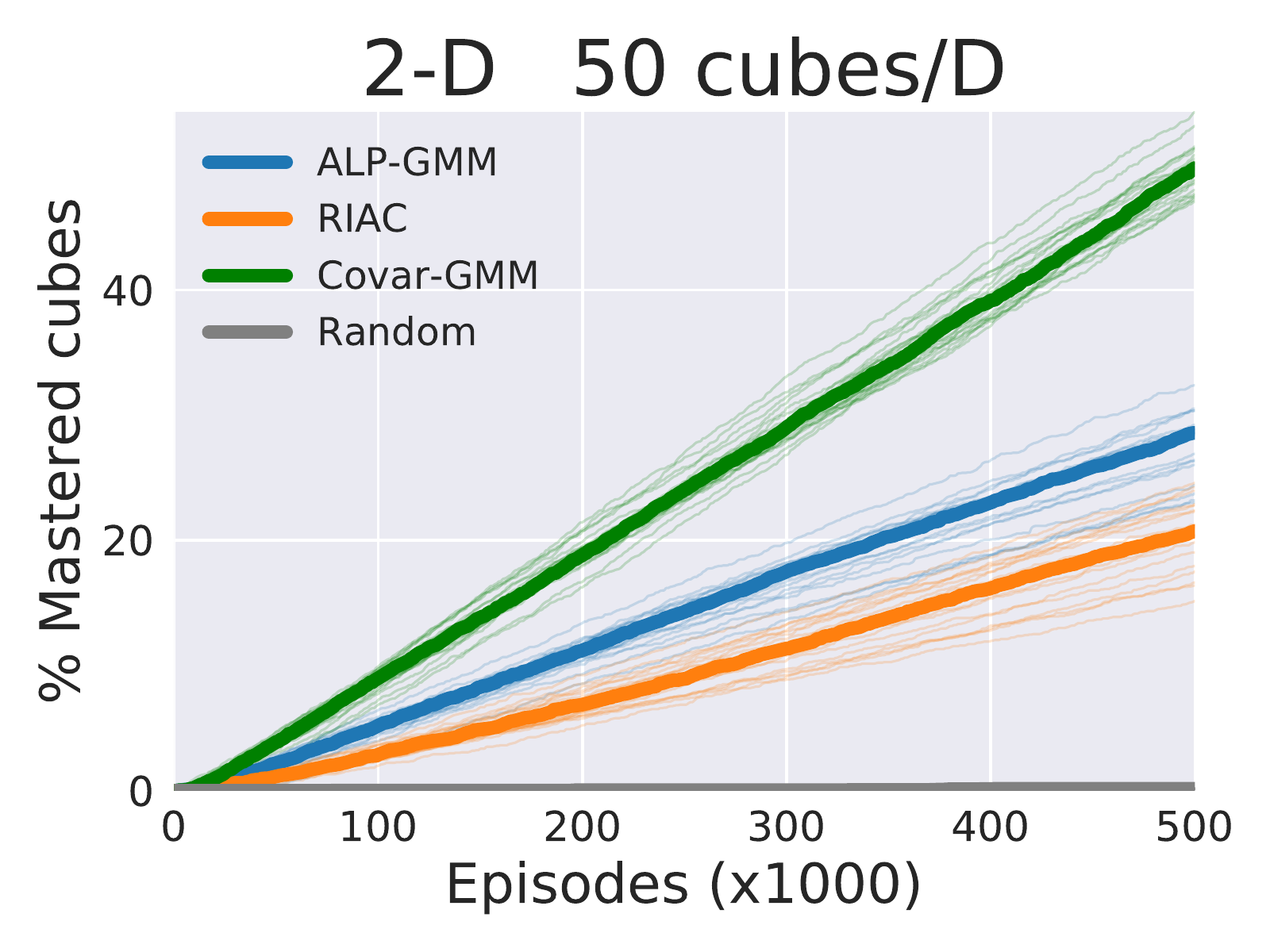}}
\subfloat{\includegraphics[width=4.65cm]{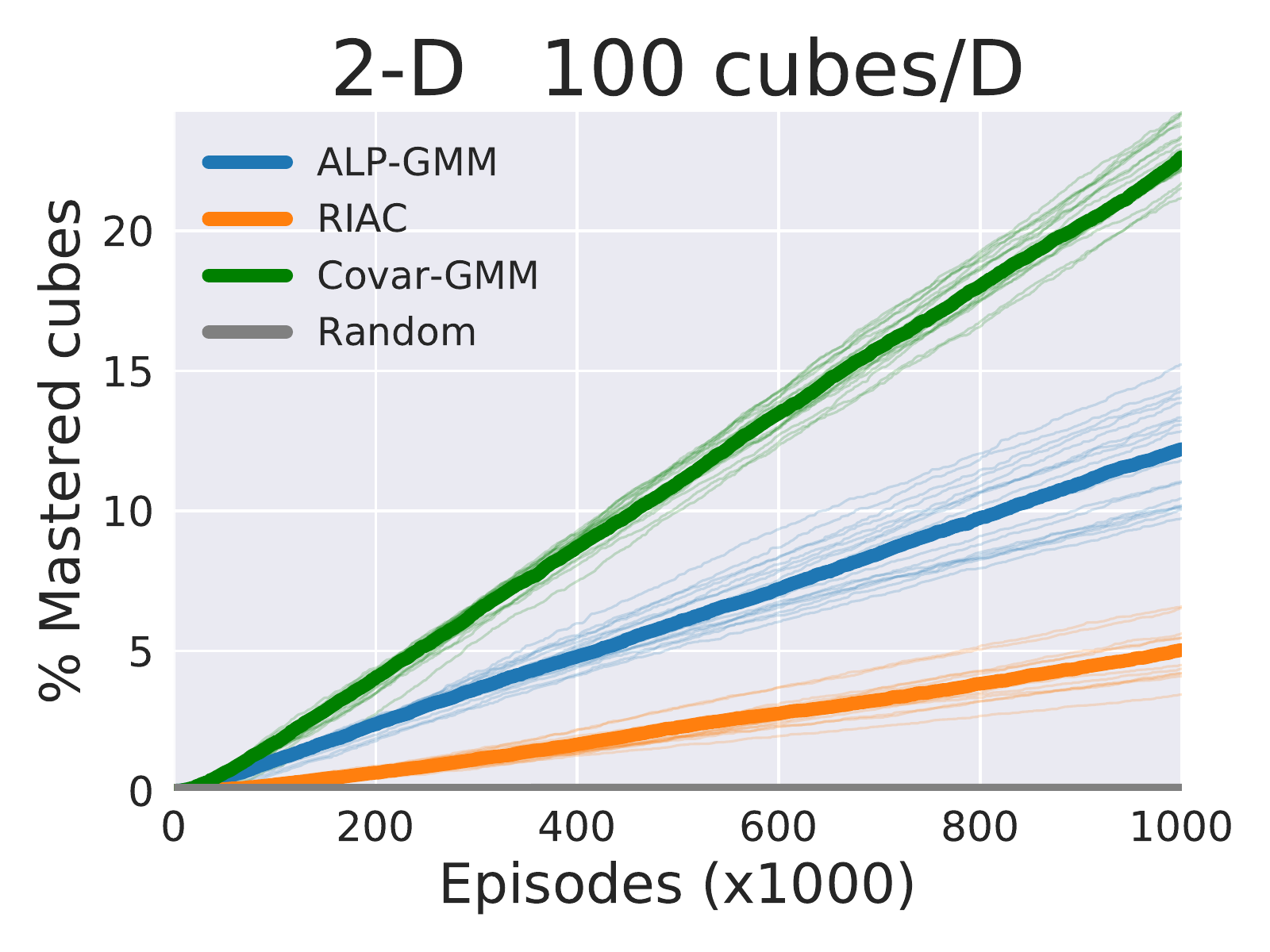}}
\caption{\footnotesize{\textbf{Evolution of performance on n-dimensional toy-spaces.} The impact of 3 aspects of the parameter space are tested: growing number of meaningful dimensions (top row), growing number of irrelevant dimensions (middle row) and increasing number of hypercubes (bottom row). The median performance (percentage of unlocked hypercubes) is plotted with shaded curves representing the performance of each run. 20 repeats were performed for each condition (for each toy-space).}}
\label{all_perfs_gridworld}
\end{figure*}

\paragraph{Results on n-dimensional toy spaces}
Results are displayed in Figure \ref{all_perfs_gridworld}. We use the median percentage of unlocked hypercubes as a performance metric. A first experiment was performed on a $2$D toy space with $10$ hypercubes per dimensions. In this experiment one can see that all LP-based approaches outperform Random by a significant margin. Covar-GMM is the highest performing algorithm. This first toy-space will be used as a point of reference for our following analysis, for which all conditions were tested on a panel of toy spaces with varying number of meaningful dimensions (first row of Figure \ref{all_perfs_gridworld}), irrelevant dimensions (second row) and number of hypercubes (third row).

By looking at the first row of Figure \ref{all_perfs_gridworld}, one can see that increasing the dimension size seems to have a greater negative impact on RIAC than on GMM-based approaches: RIAC, which was between ALP-GMM and Covar-GMM in terms of median performance in our reference experiment is now clearly under-performing them on all $3$ toy spaces. In the $3$D and $4$D cases RIAC is even outperformed by the Random condition after $70$k episodes and $290$k episodes, respectively. For the $6$D toy space RIAC consistently outperforms Random, reaching a median final performance of $80$\% after $1$M episodes. In this $6$D toy space ALP-GMM and Covar-GMM both reach $100$\% of median performance after $1$M episodes. Covar-GMM is the highest performing condition in each toy-space, closely followed by ALP-GMM.

The second row of Figure \ref{all_perfs_gridworld} shows how performances of our approaches vary when adding irrelevant dimensions to the $2$D toy-space. To better grasp the properties of these additional dimensions, one can see that Random is not affected by them. With $10$,$20$ and $50$ additional useless dimensions, RIAC is consistently inferior to GMM-based conditions in terms of median performance. RIAC median performance is only above Random during the first $55$k episodes. ALP-GMM is the highest performing algorithm throughout training for toy spaces with $20$ and $50$ irrelevant dimensions, closely followed by Covar-GMM. In the toy space with $10$ irrelevant dimensions, ALP-GMM outperforms Covar-GMM in the first $40$k episodes but end up reaching a $100$\% median performance after $52$k episodes against only $44$k episodes for Covar-GMM.

The last row shows how performance changes according to the number of hypercubes. Given our toy space rules, increasing the number of hypercubes reduces the initial area where reward is obtainable in the parameter space, and therefore allows us to study the sensitivity of our approaches to detect learnable subspaces. Random struggles in all $3$ toy-spaces compared to other conditions and to its performances on the reference experiment with $10$ hypercubes per dimensions. Covar-GMM and RIAC are the best performing conditions for toy-spaces with $20$ hypercubes per dimensions. However, when increasing to $50$ and $100$ hypercubes per dimensions, Covar-GMM remains the best performing condition but RIAC is now under-performing compared to ALP-GMM.

Overall these experiments showed that GMM-based approaches scaled better than RIAC on parameter spaces with large number of relevant or irrelevant dimensions, and large number of (initially) unfeasible parameter spaces. Among these GMM-based approaches, contrary to experiments with DRL students on BipedalWalker environments, Covar-GMM proved to be better than ALP-GMM for these toy spaces.

\clearpage

\section{Implementation details}
\label{ann:impldetails}

\paragraph{Soft-Actor Critic} All of our experiments were performed with OpenAI's implementation\footnote{\url{https://github.com/openai/spinningup}} of SAC as our DRL student. We used the same 2-layered (400, 300) network architecture with ReLU for the Q, V and policy networks. The policy network's output uses tanh activations. The entropy coefficient and learning progress were respectively set to $0.005$ and $0.001$. Gradient steps are performed every $10$ environment steps by selecting $1000$ samples from a replay buffer with a fixed sized of $2$ millions.

\paragraph{RIAC} To avoid a known tendency of RIAC to oversplit the parameter space \citep{goalgan}, we added a few modifications to the original architecture. The essence of our modified RIAC could be summarized as follows (hyperparameters settings are given in parenthesis):
\begin{enumerate}
    \item When collecting a new parameter-reward pair, it is added to its respective region. If this region reaches its maximal capacity $max_{s}$ ($=200$), a split attempt is performed.
    \item When trying to split a parent region $p$ into two children regions,  $n$ ($=50$) candidate splits on random dimensions and thresholds are generated. Splits resulting in one of the child regions, $c_1$ or $c_2$, having less than $min_s$ ($=20$) individuals are rejected. Likewise, to avoid having extremely small regions, a minimum size $min_d$ is enforced for each region's dimensions (set to $1/6$ of the initial range of each dimensions of the parameter space). The split with the highest score, defined as $card(c_1) \cdot card(c_2) \cdot |alp(c_1) - alp(c_2)|$, is kept. If no valid split was found, the region flushes its oldest points (the oldest quarter of pairs sampled in the region are removed).
    \item At sampling time, several strategies are combined:
            \begin{itemize}
                \item $20\%$: a random parameter is chosen in the entire space.
                \item  $70\%$: a region is selected proportionally to its ALP and a random parameter is sampled within the region.
                \item $10\%$: a region is selected proportionally to its ALP and the worst parameter with lowest associated episodic reward is slightly mutated (by adding a Gaussian noise $\mathcal{N}(p, 0.1)$).
            \end{itemize}
\end{enumerate}
    We send the reader back to the original papers of RIAC \citep{riac, baranes2013active} for detailed motivations and pseudo-code descriptions.
 
 \paragraph{Covar-GMM} Originating from the developmental robotic field \citep{moulinfriergmm}, this approach inspired the design of ALP-GMM. In Covar-GMM, instead of fitting a GMM on the parameter space concatenated with ALP as in ALP-GMM, they concatenate each parameters with its associated episodic return and time (relative to the current window of considered parameters). New parameters are then chosen by sampling on a Gaussian selected proportionally to its positive covariance between time and episodic reward, which emulates positive LP. Contrary to ALP-GMM, they ignore negative learning progress and do not have a way to detect long term LP (i.e LP is only measured for the currently fitted datapoints). Although not initially present in Covar-GMM, we compute the number of Gaussians online as in ALP-GMM to compare the two approaches solely on their LP measure. Likewise, Covar-GMM is given the same hyperparameters as ALP-GMM (see section \ref{sec:methods}).
 
 \paragraph{Oracle} Oracle has been manually crafted based on knowledge acquired over multiple runs of the algorithm. It uses a step size $\sigma_W = \frac{1}{30}R$, with $R$ a vector containing the maximal distance for each dimension of the parameter space. Before each new episode, the window ($W_{size} = \frac{1}{6}R$) is slid toward more complex task distributions by $\sigma_W$ only if the average episodic reward of the last $50$ proposed tasks is above $r_{thr}=230$. See Algorithm \ref{algo:linear} for pseudo-code.
 
\begin{figure*}[htb!]
\centering
\subfloat[1000 episodes]{\includegraphics[height=4.8cm]{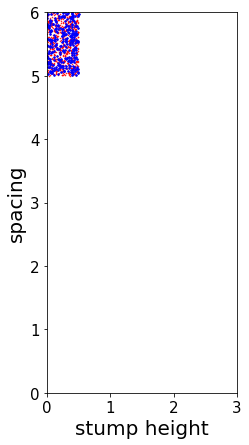}}
\subfloat[2000 episodes]{\includegraphics[height=4.8cm]{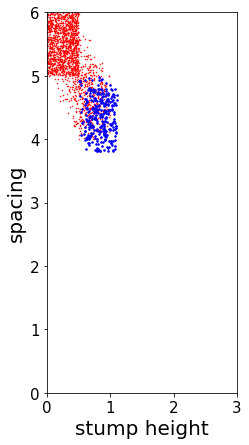}}
\subfloat[10500 episodes]{\includegraphics[height=4.8cm]{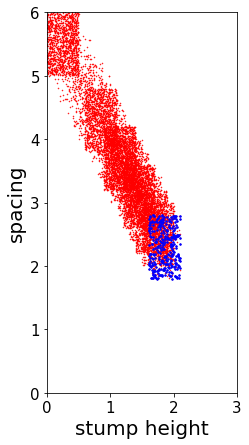}}
\subfloat[15000 episodes]{\includegraphics[height=4.8cm]{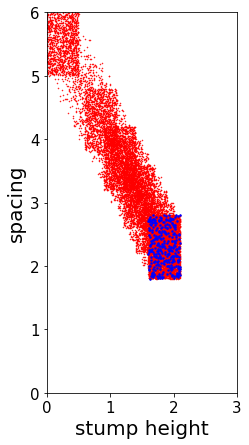}}
\caption{\footnotesize{\textbf{Evolution of Oracle parameter sampling for a default bipedal walker on Stump Tracks.} Blue dots represent the last $300$ sampled parameters, red dots represent all other previously sampled parameters. At first (a), Oracle starts by sampling parameters in the easiest subspace (i.e large stump spacing and low stump height). After $2000$ episodes (b), Oracle slid its sampling window towards stump tracks whose stump height lies between $0.6$ and $1.1$ and a spacing between  $3.7$ and $4.7$. After $10500$ episodes (c) this Oracle run reached a challenging subspace that his student will not be able to master. By $15000$ episodes, The sampling window did not move as the mean reward threshold was never crossed.}}
\label{oracle_vizu}
\end{figure*}

 Figure \ref{oracle_vizu} provides a visualization of the evolution of Oracle's parameter sampling for a typical run in Stump Tracks. One can see that the final position of the parameter sampling window (corresponding to a subspace that cannot be mastered by the student) is reached after $10500$ episodes (c) and remains the same up to the end of the run, totaling $15000$ episodes. This end-of-training focus on a specific part of the parameter space is the cause of the forgetting issues of Oracle (see section \ref{sec:results:1}).
 \begin{algorithm}[H]
	\caption{~ Oracle}
	\label{algo:linear}
	\begin{algorithmic}[1]
	
	\Require  Student $\mathcal{S}$, parametric environment $E$, bounded parameter space $\mathcal{P}$, initial sampling window position $W_{pos}$, window step size $\sigma_W$, memory size $m_{size}$, reward threshold $r_{thr}$, window-size $W_{size}$.
	\State Set sampling window $W \subset \mathcal{P}$ to $W_{pos}$
	\Loop
	\State Sample random parameter $p \in W$, send $E(\tau \sim \mathcal{T}(p))$ to $\mathcal{S}$, observe episodic reward $r_p$
	\State \textbf{If} the mean competence over the last $m_{size}$ episodes exceeds $r_{thr}$ \textbf{then} \\ \hspace{25.0pt}
	$w_{pos} = w_{pos} + \sigma_w$
	\EndLoop
	\State \textbf{Return} $\mathcal{S}$
	
	\end{algorithmic}
\end{algorithm}

\begin{figure*}
\centering
\subfloat{\includegraphics[width=\textwidth]{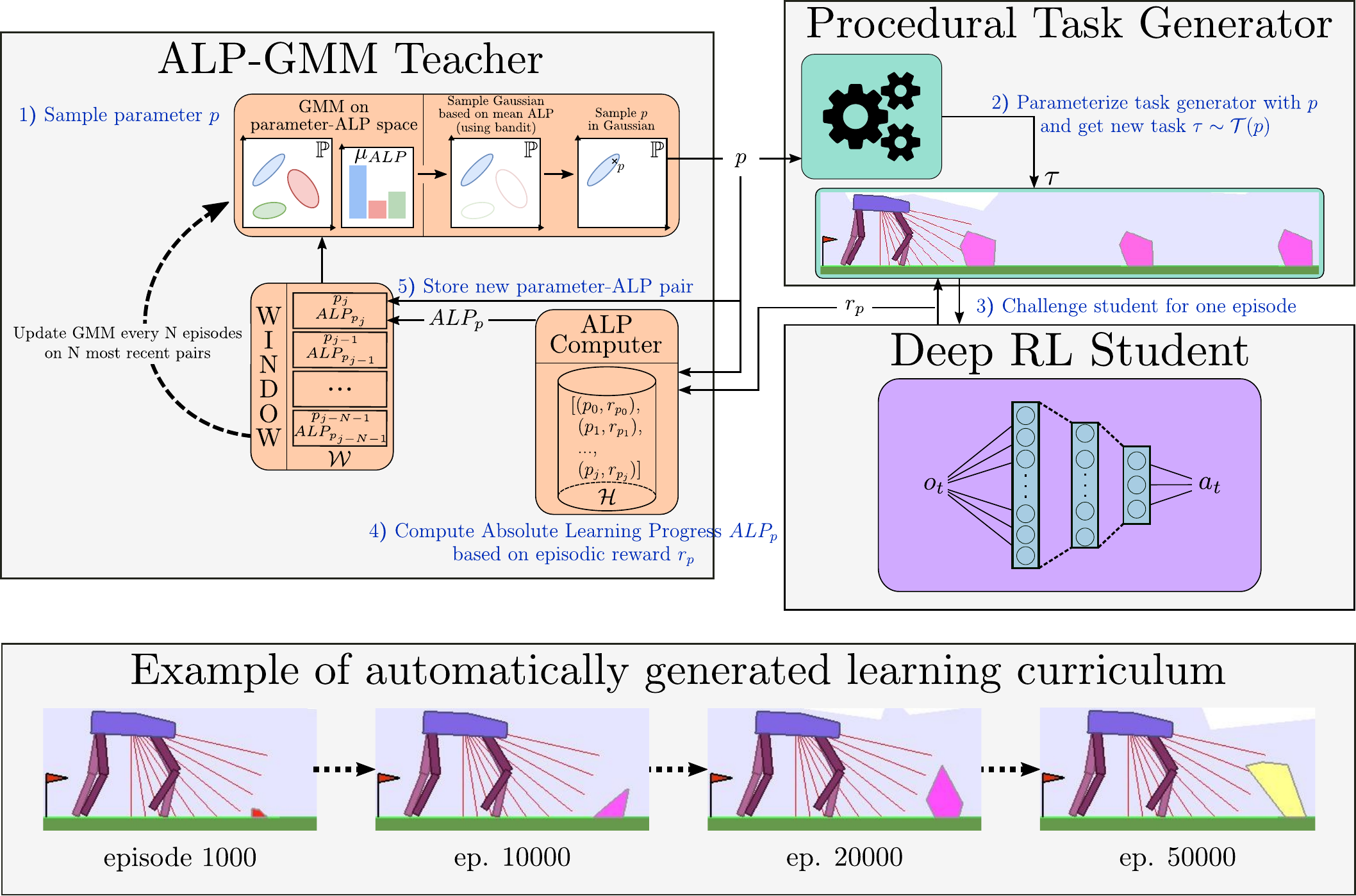}}
\caption{\footnotesize{Schematic view of an ALP-GMM teacher's workflow}}
\end{figure*}

\clearpage

\section{Additional visualizations for Stump Tracks experiments}
\label{ann:expdetails}




\begin{figure*}[htb!]
\centering
\subfloat[500 episodes]{\includegraphics[height=4.8cm]{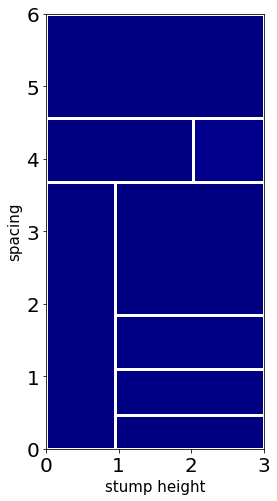}}
\subfloat[1500 episodes]{\includegraphics[height=4.8cm]{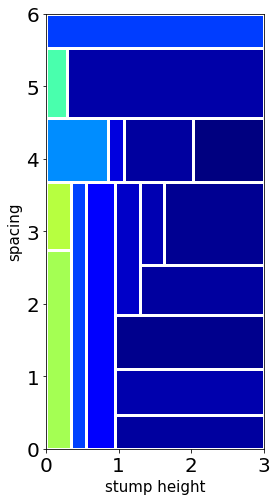}}
\subfloat[15000 episodes]{\includegraphics[height=4.8cm]{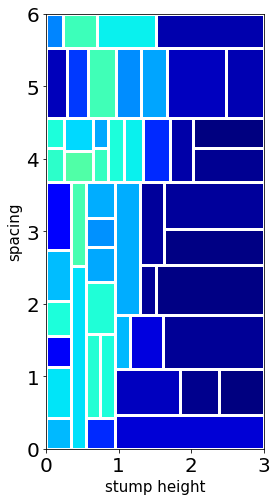}}
\subfloat[20000 episodes]{\includegraphics[height=4.8cm]{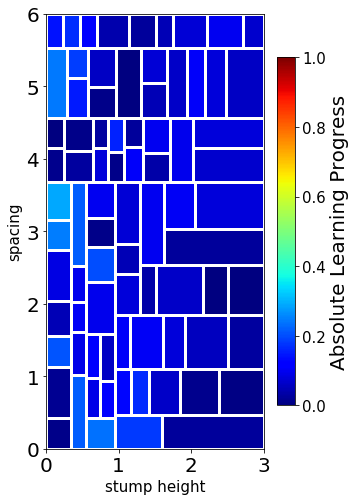}}
\caption{\footnotesize{\textbf{Evolution of RIAC parameter sampling for a default bipedal walker on Stump Tracks.} At first (a), RIAC do not find any learning progress signal in the space, resulting in random splits. After $1500$ episodes, RIAC focuses its sampling on the leftmost part of the space, corresponding to low stump heights, for which the SAC student manages to progress. After $15$k episodes (c), RIAC spreaded its sampling to parameters corresponding to track distributions with stump heights up to $1.5$, with the highest stumps paired with high spacing. By the end of the training (d) the student converged to a final skill level, and thus LP is no longer detected by RIAC, except for simple track distributions in the leftmost part of the space in which occasional forgetting of walking gates leads to ALP signal.}}
\label{riac_vizu}
\end{figure*}

\begin{figure*}[htb!]
\centering
\subfloat[Short agents]{\includegraphics[width=4.6cm]{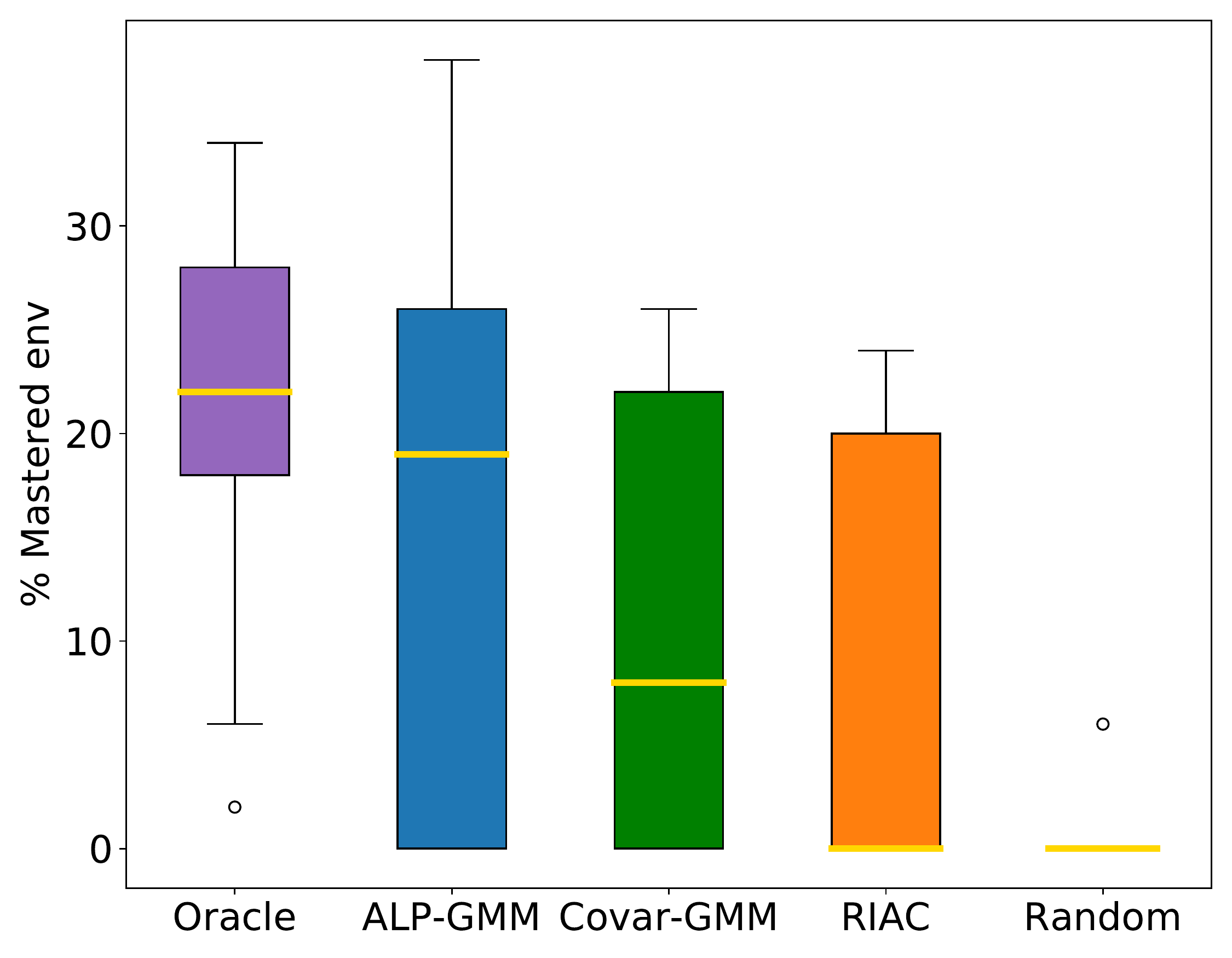}}
\subfloat[Default agents]{\includegraphics[width=4.6cm]{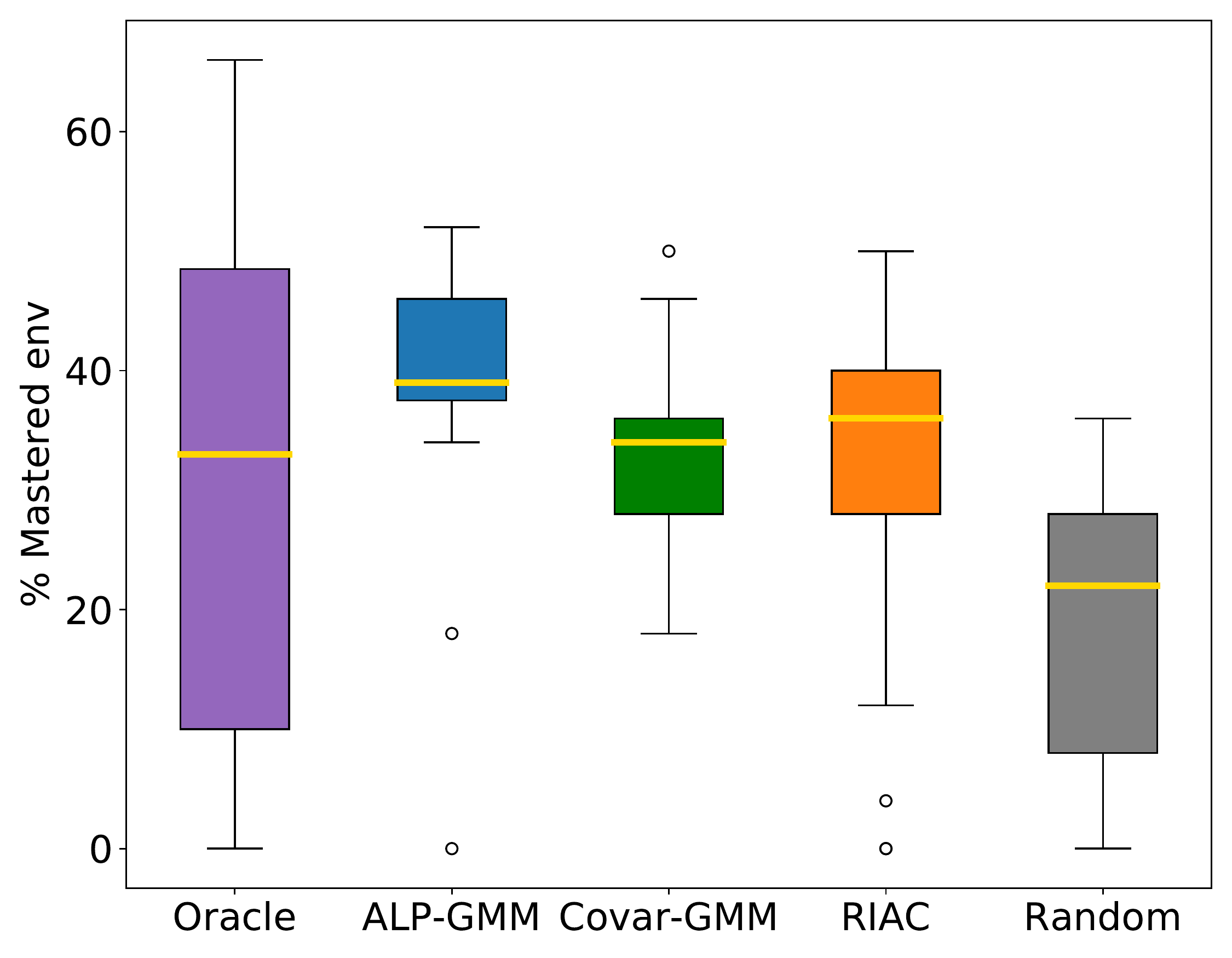}}
\subfloat[Quadrupedal agents]{\includegraphics[width=4.6cm]{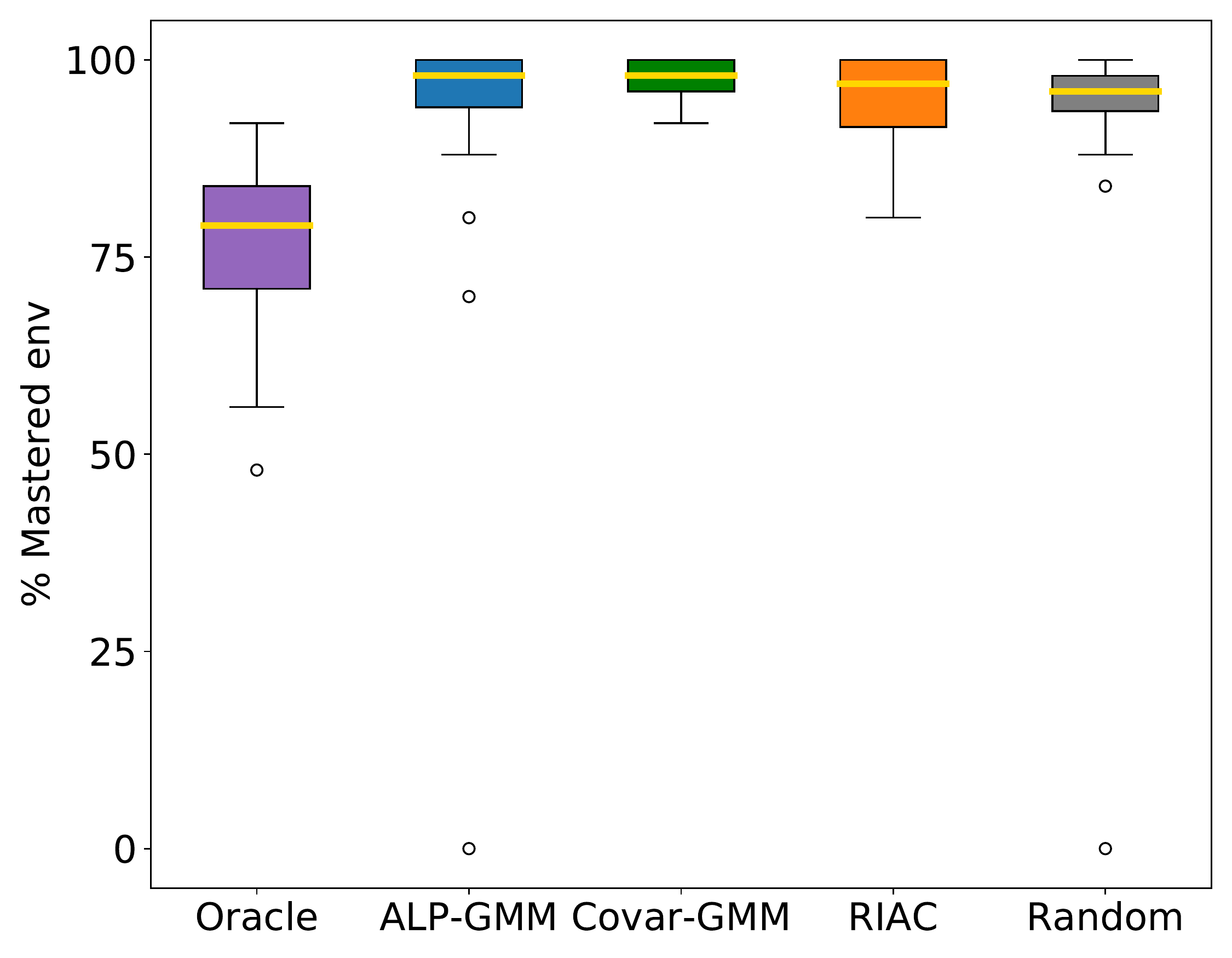}}
\caption{\footnotesize{\textbf{Box plot of the final performance of each condition on Stump Tracks after 20M steps.} Gold lines are medians, surrounded by a box showing the first and third quartile, which are then followed by whiskers extending to the last datapoint or $1.5$ times the inter-quartile range. Beyond the whiskers are outlier datapoints. \textbf{(a)}: For short agents, Random always end-up mastering 0\% of the track distributions of the test set, except for a single run that is able to master 3 track distributions (6\%). LP-based teachers obtained superior performances than Random while still failing to reach non-zero performances by the end of training in $13/32$ runs for ALP-GMM, $15/32$ for Covar-GMM and $19/32$ for RIAC. \newline \textbf{(b)}: For default walkers, LP-based approaches have less variance than Oracle (visible by the difference in inter-quartile range) whose window-sliding strategy led to catastrophic forgetting occurring in a majority of runs. Random remains the least performing algorithm. \newline \textbf{(c)}: For quadrupedal walkers, Oracle performs significantly worse than any other condition ($p<10^{-5}$). Additional investigations on the data revealed that, by sliding its sampling window towards track distributions with higher stump heights and lower stump spacing, Oracle's runs mostly failed to master track distributions that were both hard and distant from its sampling window within the parameter space: that is, tracks with both high stump heights ($>2.5$) and high spacing ($>3.0$).}}
\label{all_perfs_simple_std}
\end{figure*}

\begin{figure*}[t]
\centering
\subfloat[max stump height of $3$]{\includegraphics[width=4.65cm]{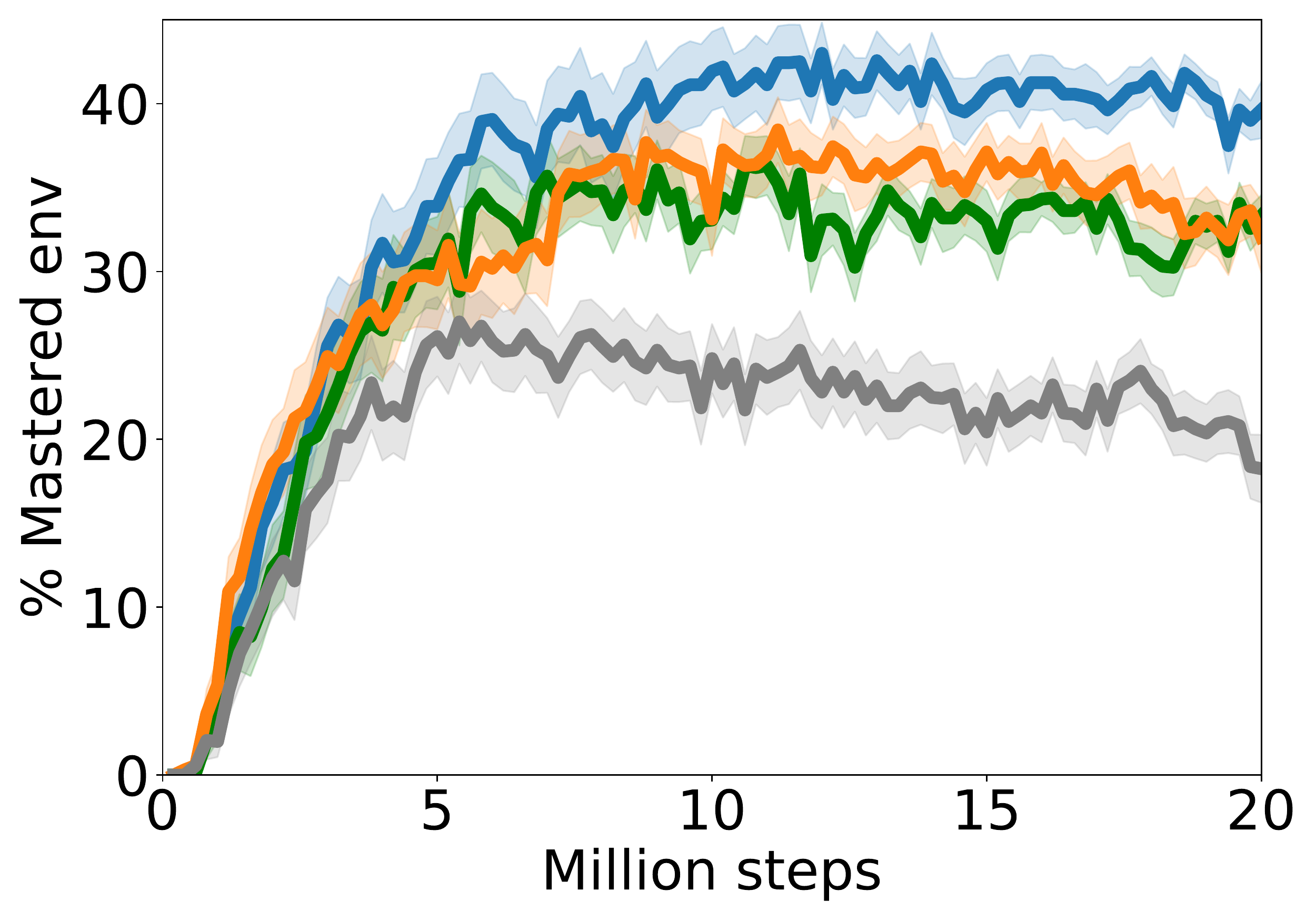}}
\subfloat[max stump height of $4$]{\includegraphics[width=4.65cm]{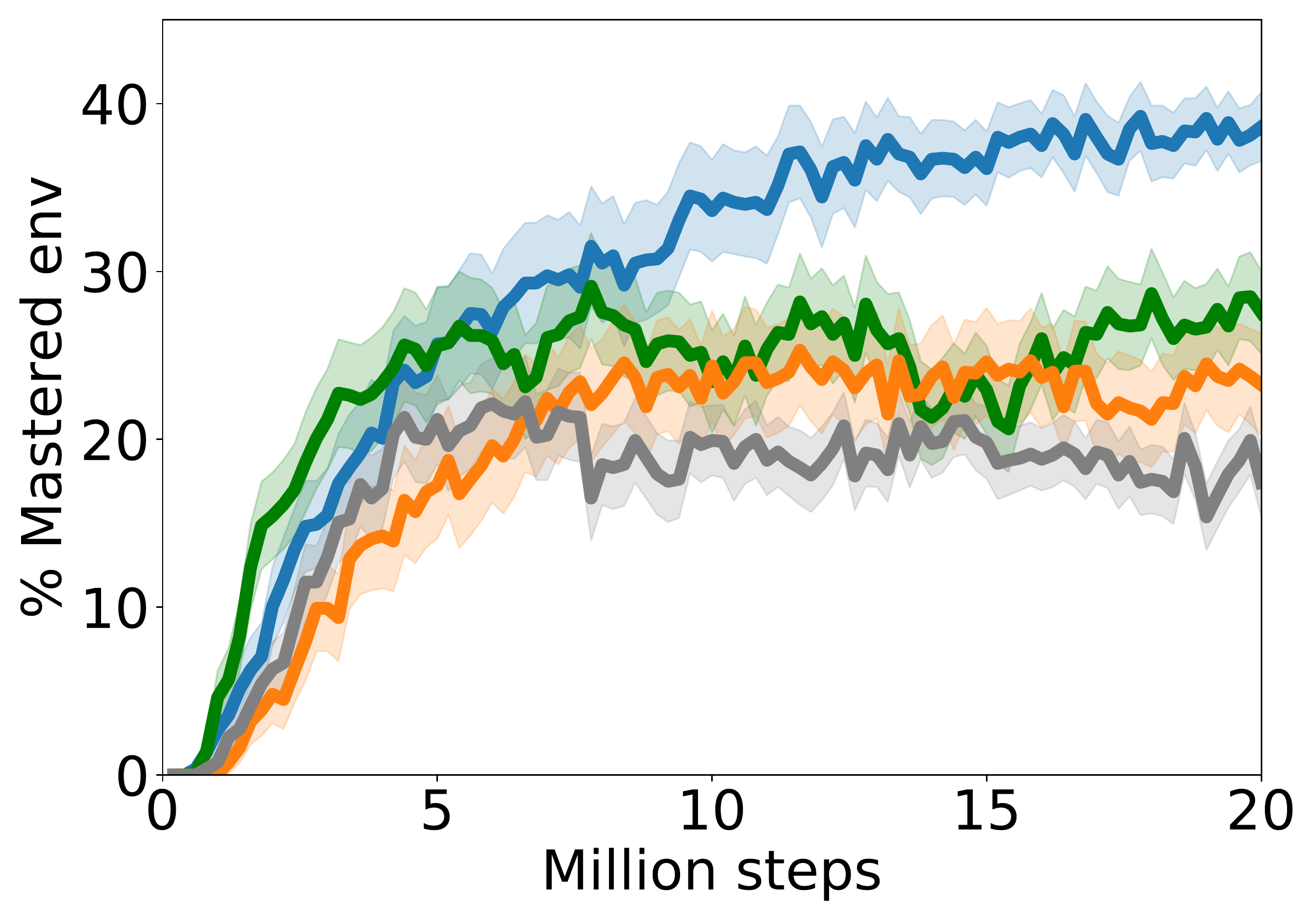}}
\subfloat[max stump height of $5$]{\includegraphics[width=4.65cm]{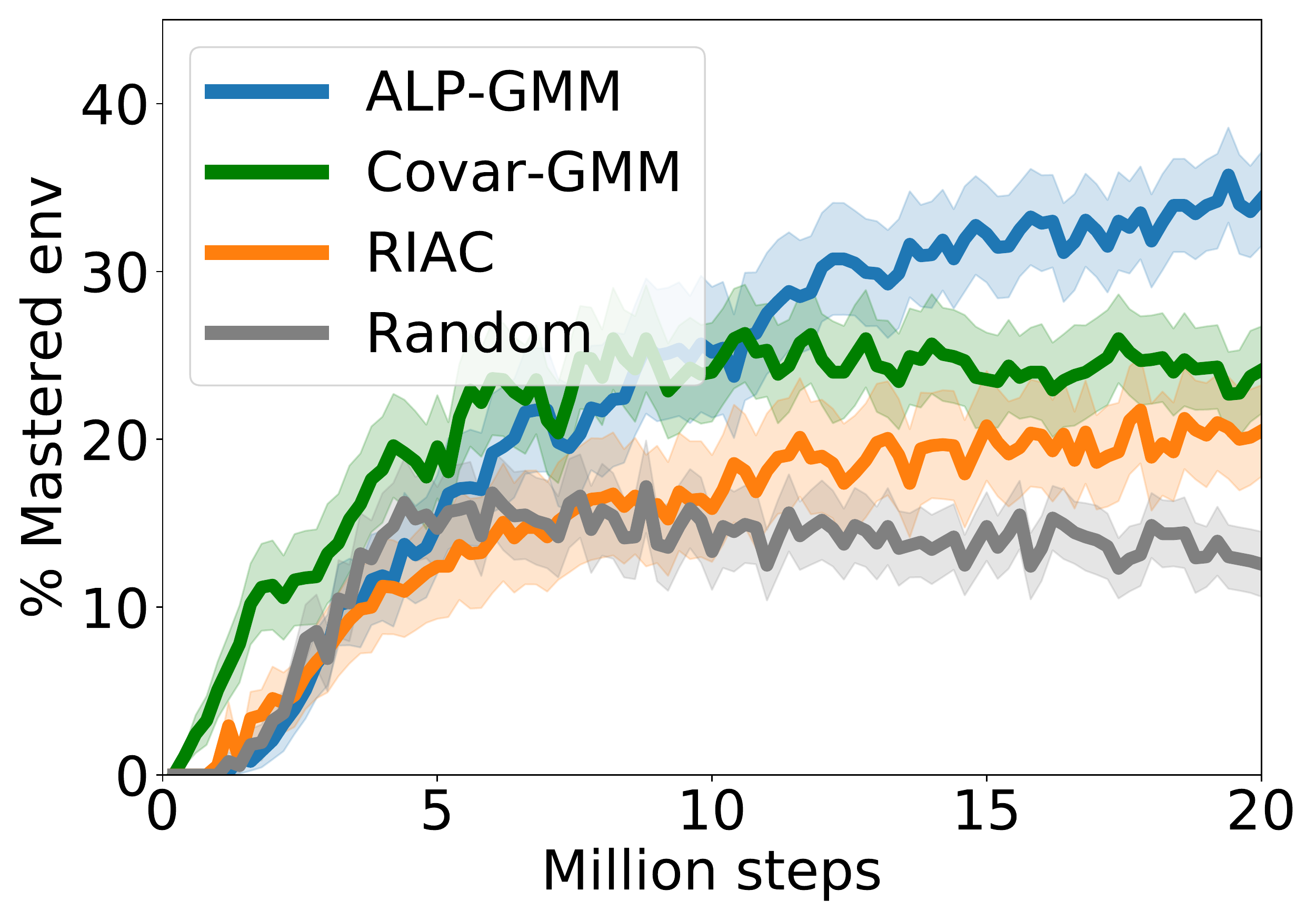}}
\caption{\footnotesize{\textbf{Evolution of mean performance of Teacher-Student approaches when increasing the amount of unfeasible tracks in Stump Tracks with default bipedal walkers.} 32 seeded runs where performed for each condition. The mean performance is plotted with shaded areas representing the standard error of the mean. ALP-GMM is the most robust LP-based teacher and maintains a statistically significant performance advantage over all other conditions in all 3 settings. Random performances are most impacted when increasing the number of unfeasible tracks. ALP-GMM is more robust than RIAC when going from a maximal stump height of $3$ to $4$ and $3$ to $5$. Note that for all 3 experiments, for comparison purposes, the same test set was used and contained only track distributions with a maximal stump height of $3$.}}
\label{all_perfs_height_exps}
\end{figure*}

\clearpage
\section{Additional visualization for Hexagon Tracks experiments}
\label{ann:expdetailshexa}

\begin{figure*}[htb!]
\centering
\subfloat{\includegraphics[width=0.7\textwidth]{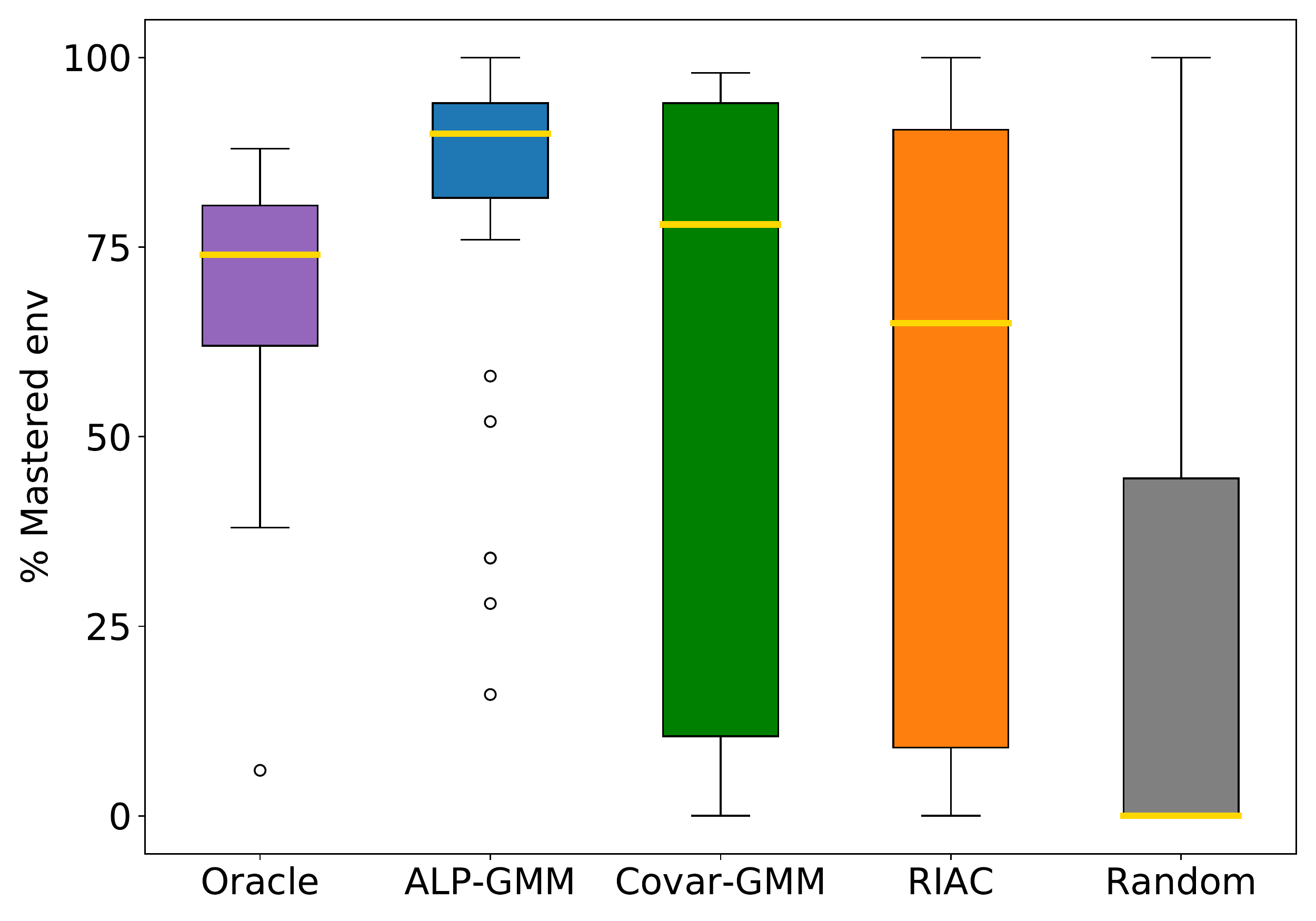}}
\caption{\footnotesize{\textbf{Box plot of the final performance of each condition run on Hexagon Tracks after 80M steps.} Gold lines are medians, surrounded by a box showing the first and third quartile, which are then followed by whiskers extending to the last datapoint or $1.5$ times the inter-quartile range. Beyond the whiskers are outlier datapoints.}}
\label{boxplot_hexa}
\end{figure*}

To better understand the properties of all of the tested conditions in Hexagon Tracks, we analyzed the distributions of the percentage of mastered environments of the test set after training for $80$ Millions (environment) steps. Using Figure \ref{boxplot_hexa}, one can see that ALP-GMM both has the highest median performance and narrowest distribution. Out of the $32$ repeats, only Oracle and ALP-GMM always end-up with positive final performance scores whereas Covar-GMM, RIAC and Random end-up with $0\%$ performance in $8/32$, $5/32$ and $16/25$ runs, respectively. Interestingly, in all repeats of any condition, the student manages to master part of the test set at some point (i.e non-zero performance), meaning that runs that end-up with $0\%$ final test performance actually experienced catastrophic forgetting. This showcase the ability of ALP-GMM to avoid this forgetting issue through efficient tracking of its student's absolute learning progress.

\clearpage
\section{Parameterized BipedalWalker Environments}
\label{app:pbw}

 In BipedalWalker environments, observations vectors provided to walkers are composed of 10 lidar sensors (providing distance measures), the hull angle and velocities (linear and angular), the angle and speed of each hip and knee joints along with a binary vector which informs whether each leg is touching the ground or not. This sums up to $24$-dimensions for our two bipedal walkers and $34$ for the quadrupedal version. To account for its increased weight and additional legs, we increased the maximal torque usage and reduced the torque penalty for quadrupedal agents.
 
\begin{figure*}[htb!]
\centering
\subfloat{\includegraphics[width=0.7\textwidth]{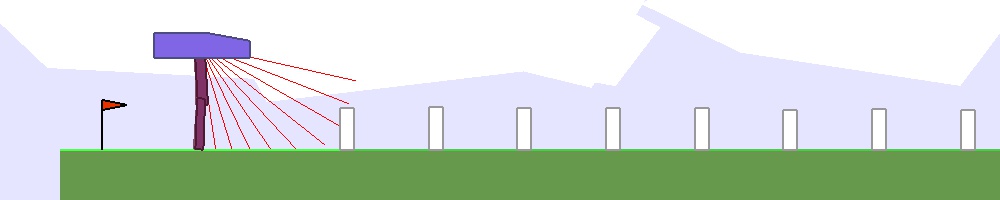}}

\subfloat{\includegraphics[width=0.7\textwidth]{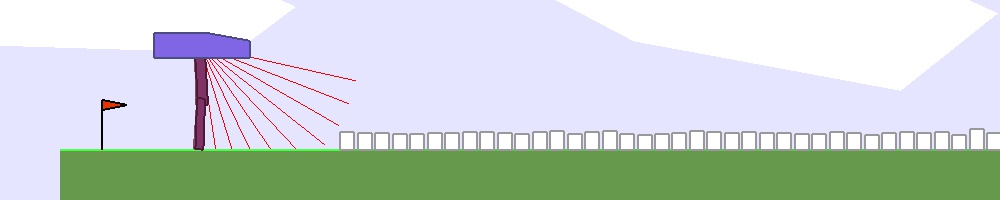}}

\subfloat{\includegraphics[width=0.7\textwidth]{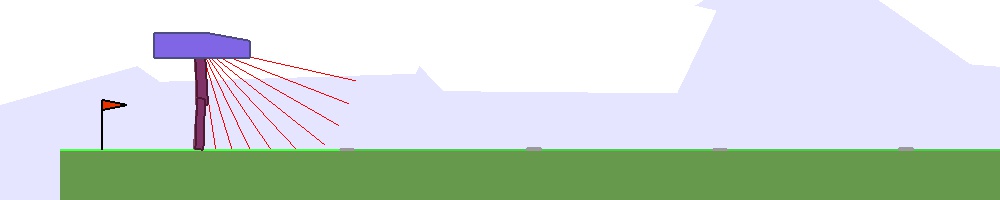}}

\subfloat{\includegraphics[width=0.7\textwidth]{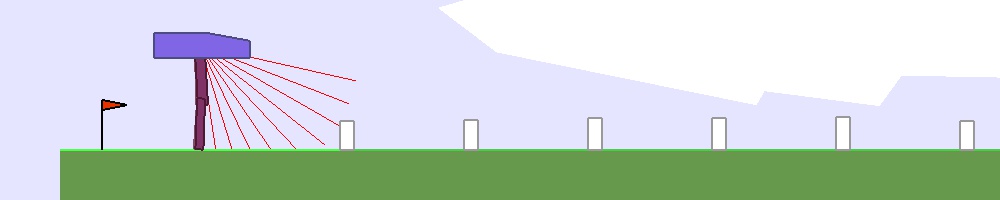}}

\subfloat{\includegraphics[width=0.7\textwidth]{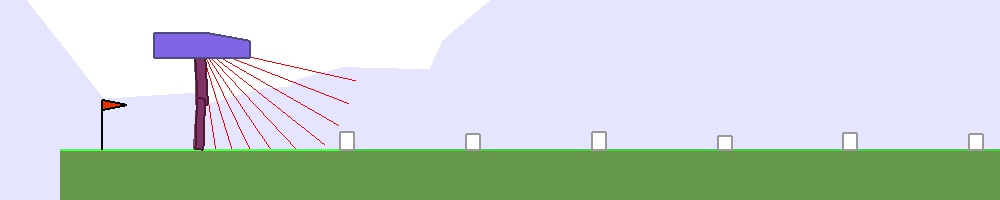}}

\caption{\footnotesize{Example of tracks generated in Stump Tracks.}}
\label{examplestump}
\end{figure*}

 \paragraph{Parameter bounds of Stump Tracks} In Stump Tracks, the range of the mean stump height $\mu_h$ is set to $[0,3]$, while the spacing $\Delta_s$ range lies in $[0,6]$. Examples of randomly generated tracks are left for consultation in Figure \ref{examplestump}.
 
\begin{figure*}
\centering
\subfloat{\includegraphics[width=0.9\textwidth]{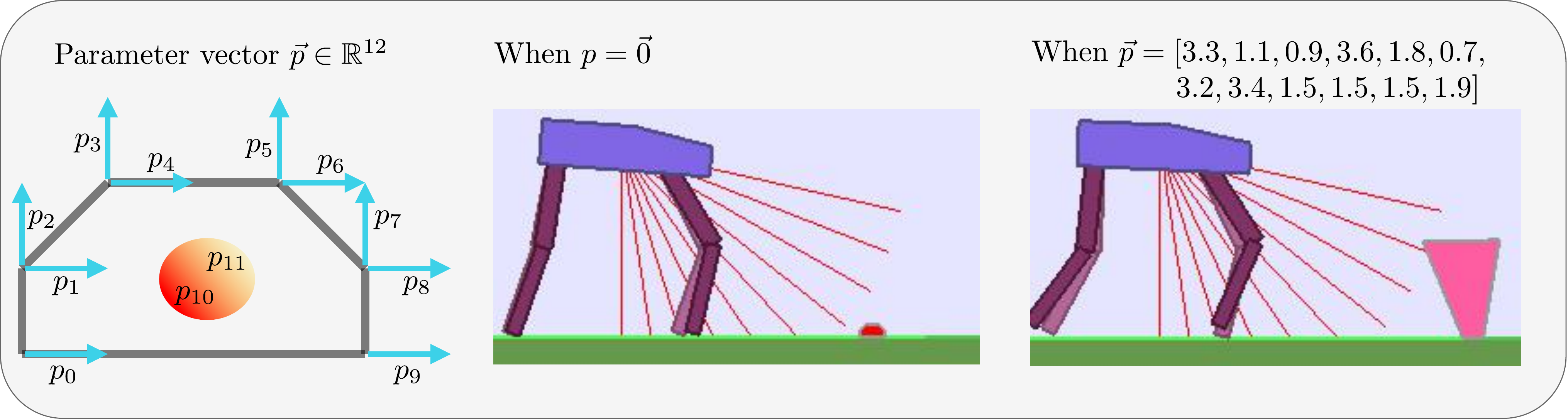}}
\caption{\footnotesize{\textbf{Generation of obstacles in Hexagon Tracks}. Given a default hexagonal obstacle, the first $10$ values of a $12$-$D$ parameter are used as positive offsets to the $x$ and $y$ positions of all vertices (except for the $y$ position of the first and last ones, in order to ensure the obstacle has at least one edge in contact with the ground.}}
\label{hexagons}
\end{figure*}

\begin{figure*}
\centering
\subfloat{\includegraphics[scale=0.5]{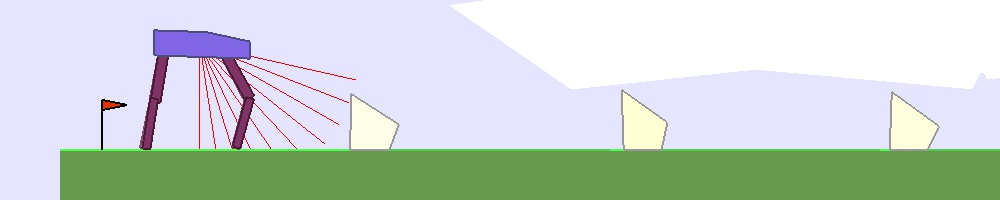}}

\subfloat{\includegraphics[scale=0.5]{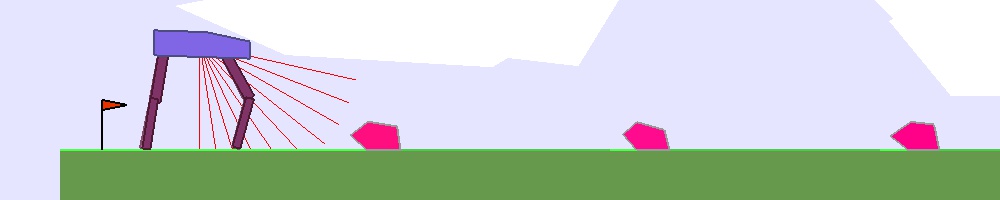}}

\subfloat{\includegraphics[scale=0.5]{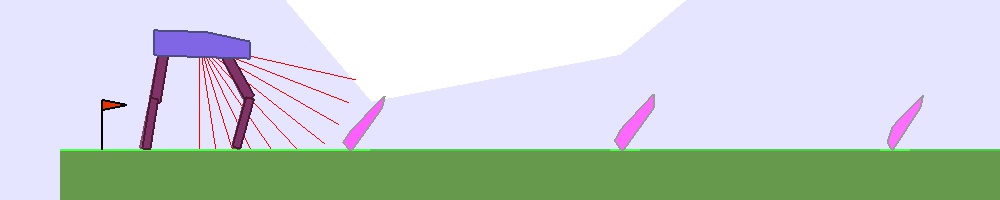}}

\subfloat{\includegraphics[scale=0.5]{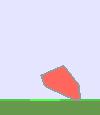}}\hspace{0.3cm}
\subfloat{\includegraphics[scale=0.5]{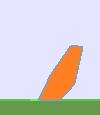}}\hspace{0.3cm}
\subfloat{\includegraphics[scale=0.5]{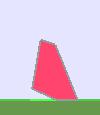}}\hspace{0.3cm}
\subfloat{\includegraphics[scale=0.5]{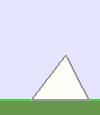}}\hspace{0.3cm}
\subfloat{\includegraphics[scale=0.5]{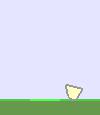}}\hspace{0.3cm}
\subfloat{\includegraphics[scale=0.5]{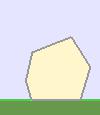}}\hspace{0.3cm}
\subfloat{\includegraphics[scale=0.5]{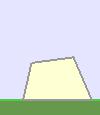}}

\subfloat{\includegraphics[scale=0.5]{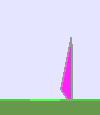}}\hspace{0.3cm}
\subfloat{\includegraphics[scale=0.5]{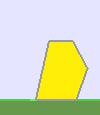}}\hspace{0.3cm}
\subfloat{\includegraphics[scale=0.5]{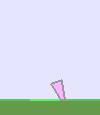}}\hspace{0.3cm}
\subfloat{\includegraphics[scale=0.5]{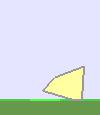}}\hspace{0.3cm}
\subfloat{\includegraphics[scale=0.5]{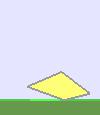}}\hspace{0.3cm}
\subfloat{\includegraphics[scale=0.5]{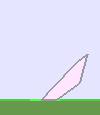}}\hspace{0.3cm}
\subfloat{\includegraphics[scale=0.5]{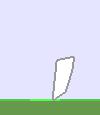}}

\caption{\footnotesize{Example of tracks generated in Hexagon Tracks with additional examples of encounterable obstacles.}}
\label{examplehexa}
\end{figure*}

 \paragraph{Parameter bounds of Hexagon Tracks} In Hexagon Tracks, the range of the $12$ dimensions of the space are set to $[0,4]$. Figure \ref{hexagons} provides a visual explanation of how hexagons are generated. Figure \ref{examplehexa} shows examples of randomly generated tracks.

\end{document}